\documentclass[journal]{IEEEtran}

\usepackage{amsmath,graphicx}
\usepackage{multicol}
\usepackage{epstopdf}
\usepackage{amsfonts,amsthm}
\usepackage{algorithm,algorithmic}
\usepackage{setspace}
\usepackage{cite}
\usepackage[colorlinks,linkcolor=blue,anchorcolor=blue,citecolor=blue]{hyperref}
\usepackage{subfloat}
\usepackage{multirow,booktabs}
\usepackage{fontenc}
\usepackage{cleveref} 

\usepackage{amsmath,amssymb,mathrsfs}
\usepackage[misc]{ifsym}
\usepackage{url}
\usepackage{bm}
\usepackage{bbding}
\usepackage{mathtools}
\usepackage{makecell}
\usepackage{subcaption}
\usepackage{hyperref}
\hypersetup{hypertex=true,
	colorlinks=true,
	linkcolor=red,
	anchorcolor=green,
	citecolor=blue}

\newtheorem{definition}{Definition}
\newtheorem{theorem}{Theorem}
\newtheorem{lemma}{Lemma}
\newtheorem{Assumption}{Assumption}
\newtheorem{remark}{Remark}

\usepackage{array}
\usepackage{cases}
\newcolumntype{"}{@{\hskip\tabcolsep\vrule width 1pt\hskip\tabcolsep}}
\usepackage{pifont}
\usepackage{color}

\ifCLASSINFOpdf
\else
\fi

\begin{document}


\title{Generalized Low-Rank Matrix Completion Model with Overlapping Group Error Representation}

\author{
Wenjing~Lu,~Zhuang~Fang,~Liang~Wu,~Liming~Tang,~Hanxin~Liu, and~Chuanjiang~He
\thanks{W. Lu(first author) is with the School of Mathematics and Statistics, Hubei Minzu University, Enshi 44500, Hubei, China (email: luwenjing07@163.com).} 
\thanks{Z. Fang (corresponding author) is with the School of Mathematics and Statistics, Hubei Minzu University, Enshi 44500, Hubei, China. (email: wdfangzhuang@163.com).}
\thanks{L. Wu is with the School of Mathematics and Statistics, Southwest University, Chongqing, 400715, China (email: liangwu1998@163.com).}
\thanks{L. Tang and H. Liu are with the School of Mathematics and Statistics, Hubei Minzu University, Enshi 44500, Hubei, China (email: tlmcs78@foxmail.com, 2531417503@qq.com).} 
\thanks{C. He is with the School of Mathematics and Statistics, Chongqing University, Chongqing, 401331, China (e-mail: chuanjianghe@sina.com).}
}

\IEEEtitleabstractindextext{%
\begin{abstract}
The low-rank matrix completion (LRMC) technology has achieved remarkable results in low-level visual tasks. There is an underlying assumption that the real-world matrix data is low-rank in LRMC. However, the real matrix data does not satisfy the strict low-rank property, which undoubtedly present serious challenges for the above-mentioned matrix recovery methods. 
Fortunately, there are feasible schemes that devise appropriate and effective priori representations for describing the intrinsic information of real data.
In this paper, we firstly model the matrix data ${\bf{Y}}$ as the sum of a low-rank approximation component $\bf{X}$ and an approximation error component $\cal{E}$. 
This finer-grained data decomposition architecture enables each component of information to be portrayed more precisely. 
Further, we design an overlapping group error representation (OGER) function to characterize the above error structure and propose a generalized low-rank matrix completion model based on OGER.
Specifically, the low-rank component describes the global structure information of matrix data, while the OGER component not only compensates for the approximation error between the low-rank component and the real data but also better captures the local block sparsity information of matrix data.
Finally, we develop an alternating direction method of multipliers (ADMM) that integrates the majorization-minimization (MM) algorithm, which enables the efficient solution of the proposed model. And we analyze the convergence of the algorithm in detail both theoretically and experimentally. In addition, the results of numerical experiments demonstrate that the proposed model outperforms existing competing models in performance.
\end{abstract}

}

\maketitle

\IEEEdisplaynontitleabstractindextext

\IEEEpeerreviewmaketitle

\section{Introduction}\label{sec:introduction}

\IEEEPARstart{L}{ow-rank} matrix completion has been extensively used in various fields as an essential technical tool for handling incomplete observation matrices, including recommender systems \cite{nie2021robust,yao2018efficient,Luo2020Non}, image restoration \cite{yang2023optimal,shi2022remove,yin2017scalable}, motion capture \cite{chen2021logarithmic,shijila2019simultaneous,peng2022exact}, and background subtraction \cite{jia2022non,zhao2022novel,yong2017robust}, etc. In practical applications, real matrix data is frequently corrupted due to unavoidable reasons, such as missing data, noise pollution, and text overwriting. 
In recent years, numerous methods have emerged that utilize low-rank priors to recover unknown low-rank matrices.
The low-rank matrix completion can be expressed as the following minimization problem,
\begin{equation}\label{eq.a1}
\begin{array}{l}
\mathop {\min }\limits_{\bf{X}} \;\;rank\left( {\bf{X}} \right)\\
s.t.\;\;\;{{\bf{Y}}_\Omega } = {{\bf{X}}_\Omega },
\end{array}
\end{equation}
where $ \Omega $ is the set of positional indices of observable matrix data $ {{\bf{Y}}_\Omega } $, $ {\bf{X}} $ is the original matrix data. 
However, the rank minimization problem is NP-hard due to the discontinuity and non-convexity of the rank function.
Therefore, finding a suitable approximation of the rank function is crucial in solving above problem \cite{recht2010guaranteed,yang2020feature,li2022matrix} .
To achieve the most accurate approximation of rank functions, nuclear norm minimization (NNM) model \cite{candes2012exact,candes2010power} have been proposed, which exploit the property of the nuclear norm as the tightest convex relaxation of the rank function and recover incomplete matrices from observation matrices containing only partial entries. 
The standard nuclear norm is defined as the sum of the singular values of matrix $ {\bf{X}} $, denoted as $ {\left\| {\bf{X}} \right\|_*} = \sum\nolimits_i {{{\left\| {{\sigma _i}\left( {\bf{X}} \right)} \right\|}_1}} $, where $ {{\sigma _i}\left( {\bf{X}} \right)} $ represents the $ i $-th singular value of matrix $ {\bf{X}} $.
However, the standard nuclear norm shrinks equally for each singular value, which results in only suboptimal performance in practical recovery applications.

Recently, several scholars have attempted to approximate the rank function utilizing non-convex functions in order to improve the data recovery performance \cite{sun2013robust,nie2012low,gu2014weighted,gu2017weighted,li2020matrix,tan2023total,wang2022low}.
In particular, Gu et al. in \cite{gu2014weighted} and \cite{gu2017weighted} considered the disparity among different singular values and introduced the weighted nuclear norm minimization (WNNM) model, which optimizes the recovery effect through adaptive weighting. 
Inspired by the WNNM model, the weighted schatten $ p $-norm was proposed in \cite{xie2016weighted}, serving as a low-rank matrix approximation method capable of capturing subtle features in matrix structures. 
Furthermore, Li et al. \cite{li2020matrix} ingeniously combined the advantages of the schatten $ p $-norm and the capped norm, devising the schatten capped $ p $ (SCP) norm, which achieves a balance between the rank and nuclear norm of the matrix.
Subsequently, the introduction of an adaptive weight function strategy \cite{zhao2024adaptive} in low-rank matrix/tensor complementation provides a more flexible method for adaptively assigning singular value weights. 
Gao et al. \cite{gao2024low} proposed a novel non-convex surrogate of the rank function, utilizing the ratio of the nuclear norm to the frobenius norm. 
These models based on non-convex surrogate functions have achieved remarkable recovery effects in practical applications, owing to their ability to accurately portray the complex properties of matrix data.

Although the above models have indeed improved the performance of data restoration to some extent, there remains a pressing issue that demands further exploration. 
That is, these models often assume that the matrix data possesses strict low-rank properties and relies solely on low-rank representations to characterize the essential features of the data in modeling \cite{candes2011robust,lin2024tensor}. 
At present, robust principal component analysis (RPCA) \cite{wright2009robust,liu2012robust}, is a widely adopted decomposition method, which decomposes the real matrix data ${\bf{Y}}$ into a low-rank component $\bf{X}$ and a sparse component ${\bf{S}}$. It models the decomposition as follows,
\begin{equation}\label{eq.a2}
\begin{array}{l}
\mathop {\min }\limits_{\bf{X}} \;rank\left( {\bf{X}} \right){\rm{ + }}\lambda {\left\| {\bf{S}} \right\|_0}\\
s.t.\;\;\;{\bf{Y}} = {\bf{X}} + {\bf{S}},
\end{array}
\end{equation}
where $ \bf{S} $ is employed to capture noise, outliers, or corrupted portions of the  matrix data, and $\lambda$ is a positive regularization parameter.
Similar to problem \eqref{eq.a1}, problem \eqref{eq.a2} is also an NP-hard problem due to the non-convexity of the rank function and the $ {L^0} $-norm. To facilitate handling this problem, the nuclear norm and $ {L^1} $-norm were introduced in \cite{zhou2010stable} as convex approximations to the rank function and $ {L^0} $-norm in problem \eqref{eq.a2}, respectively.

Recent studies have shown that non-convex functions often exhibit superior performance over convex functions in addressing the optimization problem of rank function relaxation. For instance, a novel non-convex penalty sparse component method was developed in \cite{chartrand2012nonconvex}, which achieves effective minimization via generalized shrinkage. The application of non-convex total variation regularization to sparse matrices in \cite{wang2021surface} enhances matrix sparsity in the gradient domain, thereby improving solution accuracy. To further preserve more information within matrix data, a new non-convex sparse function was constructed in \cite{li2020adaptive} to replace traditional sparsity constraints. Considering the effectiveness of minimally-concave penalty (MCP) as a sparsity-inducing regularization, Pokala et al.  \cite{pokala2021iteratively} proposed a method that can accurately achieve low-rank sparse matrix factorization. Especially, a solvable algorithm targeting the $ L^0 $ norm is proposed \cite{yuan2019}, albeit at the cost of increased computational complexity. It is worth noting that the non-convex relaxation method mentioned above mainly focuses on point sparse modeling, which may fall short of capturing more complex structural features present in matrix data.

\begin{figure}[!t]
	\centering
	\includegraphics[width=3.5in]{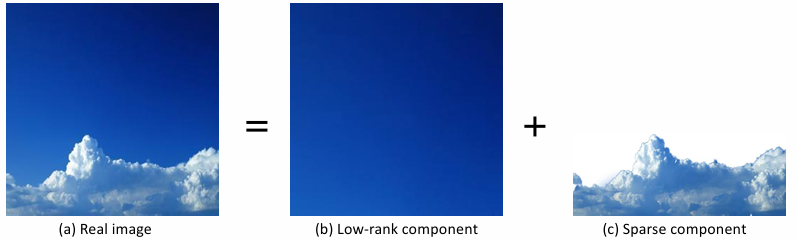}
	\caption{An example of the decomposition result of a real image.}\label{fig.1}
	\vspace{-0.5cm}
\end{figure}

However, there are two main shortcomings of the above methods. On the one hand, most of the methods mentioned above inherently assume that the sparse components decomposed from modeling are solely noise, overlooking the sparse detail information intrinsic to original data. On the other hand, these methods do not focus on the description of sparse data for structural blocks. Therefore, there is a large approximation error between the reconstructed low-rank matrix data and the original matrix data. In fact, the real data is not merely comprised of low-rank matrix, but is composed of both low-rank component and structured sparse component, as shown in Fig. \ref{fig.1}. Thus, it is a critical challenge to design an appropriate and accurate prior representation function for capturing structured sparse detail elements of real data.

\begin{figure*}[!htbp]
	\centering
	\includegraphics[width=6in]{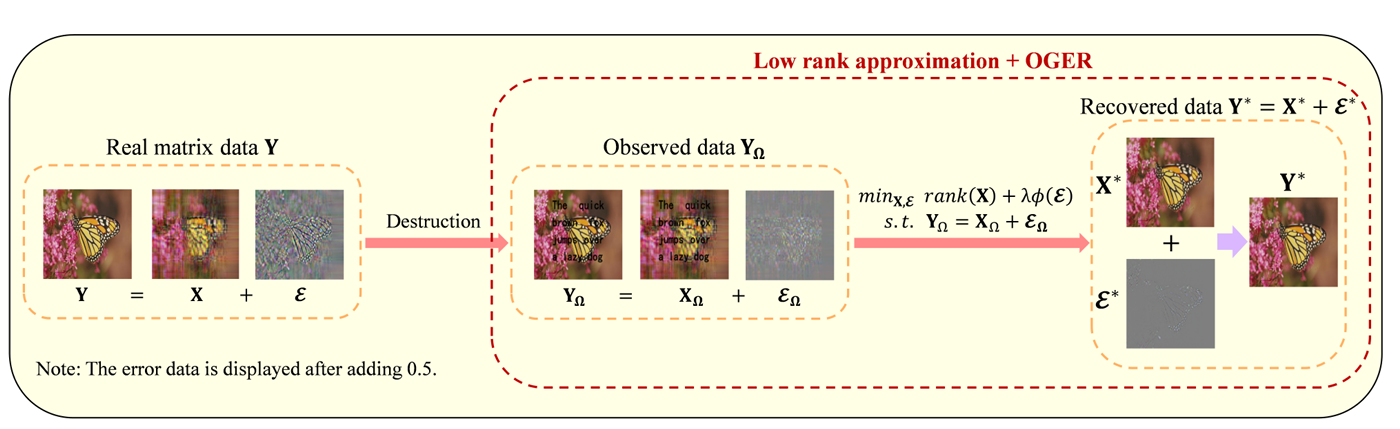}
	\caption{Illustration of the proposed method.}\label{fig.0}
	\vspace{-0.5cm}
\end{figure*}

\subsection{Our Contributions}

In this paper, we propose a generalized low-rank matrix completion model with overlapping group error representation (OGER). And the contributions of this paper are concluded as follows:

1) First, we design a priori operator for inscribing the matrix with structured sparse properties, which is called the OGER function in this paper (see Definition \ref{deoger}). On this basis, a novel decomposition framework of the real low-rank matrix data $\bf{Y}$ is proposed as
\[{{\bf{Y}} = {\bf{X}} + {\cal E}},\]
where $ {\bf{X}} $ and  $ {\cal E} $ are the low-rank component and the structured sparse component, respectively. This novel form of data decomposition effectively decomposes a complex data matrix into two components with distinct characteristics, which provides accurate modeling for further operations.

2) Second, based on the above decomposition framework, we propose a mathematical model for low-rank matrix completion (more detailed technical steps please see Fig. \ref{fig.0}), which is described as
\[
\begin{array}{l}
\mathop {\min }\limits_{{\bf{X}},{\bf{\cal E }}} R\left( {\bf{X}} \right) + \lambda \phi \left( {\bf{\cal E }} \right)\\
s.t.\;{{\bf{Y}}_\Omega }{\rm{ = }}{{\bf{X}}_\Omega }{\rm{ + }}{{\bf{\cal E }}_\Omega },
\end{array}
\]
where $\lambda>0$ is a regularization parameter, $ \Omega $ is the set of positional indices of observable matrix data $ {{\bf{Y}}_\Omega } $, $ {\bf{X}} $ and $ {\bf{\cal E }} $ are the low-rank component and structured sparse component, respectively. By minimizing the above model, it is possible to efficiently obtain the low-rank part $\bf{X}$ and the structured sparse part $\bf{\cal{E}}$ from the observed matrix ${{\bf{Y}}_\Omega }$ with missing values. Further, the recovered data $\bf{Y}^*$ is mathematically represented as the sum of these two matrix components, i.e., ${\bf{Y}^*}={\bf{X}}+{\bf{\cal E}}$.

3) Third, we propose an efficient alternating direction method of multipliers (ADMM) \cite{wei2012distributed,boyd2011distributed,lu2024coupled} that integrates the majorization-minimization (MM) algorithm \cite{hunter2004tutorial,pokala2021iteratively,wu2022hybrid} to solve the above proposed model. We analyze the proposed model and algorithm, and notice that they perform better mathematical properties including convexity of OGER (Section \ref{sec.3.1}), convergence of the algorithm (Section \ref{sec.4.5}) and so on. In addition, numerous numerical experimental results show that the proposed model and algorithm achieve excellent performance in matrix completion. Compared with other state-of-the-art models, our method still has a large advantage.

\subsection{Organization}
The remainder of this paper is structured as follows. Section \ref{sec.2} presents definitions and theorems pertinent to low-rank matrix completion methods. Section \ref{sec.3} provides a detailed exposition of the proposed model and its corresponding solution algorithm. The numerical experimental results of the model and a discussion of its parameters are shown in Section \ref{sec.4}. Finally, Section \ref{sec.5} summarizes the entire manuscript.

\section{Preliminaries and problem formulation}\label{sec.2}
The low-rank matrix completion methods have been extensively studied owing to their widespread applications in computer vision and machine learning \cite{zhang2023new,zhang2018multi}. In this section, we mainly review some definitions and theorems related to low-rank matrix completion methods.
\begin{definition}[SVD \cite{Candes2009}]\label{de.1}
	For any given matrix $ {\bf{M}} \in\mathbb{R}^{n_1\times n_2} $, its singular value decomposition is $ {\bf{M}} = {{\bf{U}}_{\bf{M}}}{{\bf{\Sigma }}_{\bf{M}}}{\bf{V}}_{\bf{M}}^{{T}} $, where $ {{\bf{U}}_{\bf{M}}}\in\mathbb{R}^{n_1\times n_1} $,	$ {{\bf{V}}_{\bf{M}}} \in\mathbb{R}^{n_2\times n_2} $ are both orthogonal matrices, $ {{\bf{\Sigma }}_{\bf{M}}} \in\mathbb{R}^{n_1\times n_2} $ is a diagonal matrix, and the elements $ {\sigma _1} \ldots {\sigma _r} $ on its diagonal are the singular values of $ \bf{M} $ and
\[{{\bf{\Sigma }}_{\bf{M}}} = \left[ {\begin{array}{*{20}{c}}
	{{\sigma _1}}&{}&{}\\
	{}& \ddots &{}\\
	{}&{}&{{\sigma _r}}
	\end{array}} \right].\]
\end{definition}

\begin{definition}[Rank Function \cite{Candes2009}]\label{de.2}
	For a given matrix $ {\bf{M}} \in\mathbb{R}^{n_1\times n_2} $ satisfying the SVD, if $ r = \min \left\{ {{n_1},{n_2}} \right\} $ and the singular values of matrix $ \bf{M} $ are $ {\sigma _1} \ge {\sigma _2} \ge {\sigma _3} \ge  \cdots  \ge {\sigma _r} $, then the rank function of matrix $ \bf{M} $ can be defined as the number of nonzero singular values in $ \bf{M} $ can be expressed as follows,
	\[rank\left( {\bf{M}} \right) = {\left\| {{{\bf{\Sigma }}_{\bf{M}}}} \right\|_0} = \# \left\{ {i|{\sigma _i}\left( {\bf{M}} \right) \ne 0} \right\}.\]
\end{definition}

\begin{definition}[Nuclear Norm \cite{candes2012exact,candes2010power}]\label{de.3}
	For a given matrix $ {\bf{M}} \in\mathbb{R}^{n_1\times n_2} $ satisfying the SVD, if $ r = \min \left\{ {{n_1},{n_2}} \right\} $ and the singular values of matrix $ \bf{M} $ are $ {\sigma _1} \ge {\sigma _2} \ge {\sigma _3} \ge  \cdots  \ge {\sigma _r} $, then the nuclear norm of matrix $ \bf{M} $ can be expressed as the following form,
\[{\left\| {\bf{M}} \right\|_*} = \sum\nolimits_{i = 1}^{\min \left\{ {{n_1},{n_2}} \right\}} {{\sigma _i}\left( {\bf{M}} \right)}. \]
\end{definition}

\begin{definition}[Schatten $ p $ Norm \cite{nie2012low}]\label{de.4}
	For a given matrix $ {\bf{M}} \in\mathbb{R}^{n_1\times n_2} $ satisfying the SVD, if $ r = \min \left\{ {{n_1},{n_2}} \right\} $ and the singular values of matrix $ \bf{M} $ are $ {\sigma _1} \ge {\sigma _2} \ge {\sigma _3} \ge  \cdots  \ge {\sigma _r} $, then the schatten $ p $ norm of matrix $ \bf{M} $ can be expressed as the following form,
	\[{\left\| {\bf{M}} \right\|_{Sp}} = {\left( {\sum\nolimits_{i = 1}^{\min \left\{ {{n_1},{n_2}} \right\}} {{\sigma _i}{{\left( {\bf{M}} \right)}^p}} } \right)^{1/p}},\]
	where $0 < p \le 1$ is a scale parameter.
\end{definition}

\begin{definition}[Weighted Nuclear Norm \cite{gu2014weighted,gu2017weighted}]\label{de.5}
	For a given matrix $ {\bf{M}} \in\mathbb{R}^{n_1\times n_2} $ satisfying the SVD, if $ r = \min \left\{ {{n_1},{n_2}} \right\} $ and the singular values of matrix $ \bf{M} $ are $ {\sigma _1} \ge {\sigma _2} \ge {\sigma _3} \ge  \cdots  \ge {\sigma _r} $, then the weighted nuclear norm of matrix $ \bf{M} $ can be expressed as the following form,
	\[{\left\| {\bf{M}} \right\|_{{\bf{w}},*}} = {\sum\nolimits_{i = 1}^{\min \left\{ {{n_1},{n_2}} \right\}} {\left\| {{w_i}{\sigma _i}\left( {\bf{M}} \right)} \right\|} _1},\]
    where $ {w_i} \ge 0 $ are weight coefficients.
\end{definition}

\begin{definition}[Weighted Schatten $ p $ Norm \cite{xie2016weighted}]\label{de.6}
	For a given matrix $ {\bf{M}} \in\mathbb{R}^{n_1\times n_2} $ satisfying the SVD, if $ r = \min \left\{ {{n_1},{n_2}} \right\} $ and the singular values of matrix $ \bf{M} $ are $ {\sigma _1} \ge {\sigma _2} \ge {\sigma _3} \ge  \cdots  \ge {\sigma _r} $, then the weighted schatten $ p $ norm of matrix $ \bf{M} $ can be expressed as the following form,
\[{\left\| {\bf{M}} \right\|_{{\bf{w}},Sp}} = {\left( {\sum\nolimits_{i = 1}^{\min \left\{ {{n_1},{n_2}} \right\}} {{w_i}{\sigma _i}{{\left( {\bf{M}} \right)}^p}} } \right)^{1/p}},\]
    where $ {w_i} \ge 0 $ and $0 < p \le 1$ are weight coefficients and scale parameter, respectively.
\end{definition}

\begin{definition}[Schatten Capped $ p $ Norm \cite{li2020matrix}]\label{de.7}
	For a given matrix $ {\bf{M}} \in\mathbb{R}^{n_1\times n_2} $ satisfying the SVD, if $ r = \min \left\{ {{n_1},{n_2}} \right\} $ and the singular values of matrix $ \bf{M} $ are $ {\sigma _1} \ge {\sigma _2} \ge {\sigma _3} \ge  \cdots  \ge {\sigma _r} $, then the schatten capped $ p $ norm of matrix $ \bf{M} $ can be expressed as the following form,
\[{\left\| {\bf{M}} \right\|_{Sp,\tau }} = {\left( {\sum\nolimits_{i = 1}^{\min \left\{ {{n_1},{n_2}} \right\}} {\min {{\left( {{\sigma _i}\left( {\bf{M}} \right),\tau } \right)}^p}} } \right)^{1/p}},\]
where $ {\tau} \ge 0 $ and $0 < p \le 1$ are shrinkage threshold and scale parameter, respectively.
\end{definition}

\begin{theorem}[MM Algorithm \cite{hunter2004tutorial,pokala2021iteratively,wu2022hybrid}]\label{de.8}
	If $ f\left( u \right) $ is a smooth function, consider the optimization problem,
	\[{u^*} = \mathop {{\rm{argmin}}}\limits_u f\left( u \right)\]
	the solution of the following iterative optimization problem converges to the solution $ {u^*} $ of the above minimization problem as $ n \to \infty $, 
	\[{u_{n + 1}} = \mathop {{\rm{argmin}}}\limits_u g\left( {u,{u_n}} \right),\]
	where the surrogate function $g\left( {u,{u_n}} \right) $ satisfies: 
	(i) For any $ u $, $ g\left( {{u_n},{u_n}} \right) = f\left( {{u_n}} \right) $ and (ii) $ g\left( {u,{u_n}} \right) \ge f\left( u \right) $. The detailed optimization steps are shown in Algorithm \ref{alg.0}.
\end{theorem}

\begin{algorithm}[tbp]
	\setstretch{1}
	\renewcommand{\algorithmicrequire}{ \textbf{Input}:}
	\renewcommand{\algorithmicensure}{ \textbf{Output}:}
	\caption{MM algorithm} \label{alg.0}
	\begin{algorithmic}[1]
		\REQUIRE $ g\left(  \cdot  \right)$, $N$.
		\STATE Initialize $ u$ and $n=0$.
		\STATE \textbf{while} not converged \textbf{do}
		\STATE \quad Update $u_{n+1}$ by ${u_{n + 1}} = \arg {\min _u}g\left( {u,{u_n}} \right)$;
		\STATE \quad $n=n+1$.
		\STATE \textbf{end while}
		\vspace{0.3em}
		\ENSURE ${u^*}=u_{n+1}$.
	\end{algorithmic}
\end{algorithm}

\begin{remark}\label{re.1}
	When the original objective function in an optimization problem has a complex structure, we usually seek an surrogate function that satisfies the above two preconditions to facilitate the simplification of the form of the function. In general, the surrogate function is convex in nature. In this case, it is pointed out in the literature \cite{sun2016majorization} that if the original objective function $ f\left( u \right) $ is convex, as $ n \to \infty $, the solution $ {u_{n + 1}} $ obtained by the MM algorithm converges to the original actual minimum point $ {u^*} $; if the original objective function $ f\left( u \right) $ is nonconvex, as $ n \to \infty $, the solution $ {u_{n + 1}} $ obtained by the MM algorithm converges to the extreme value point (also called the stabilization point) of the original objective function.
\end{remark}

\section{Establishment of the model}\label{sec.3}
As mentioned in the introduction section, most real data are not strictly low-rank, and thus it is not accurate to portray their essential features solely through low-rank representations. 
In this section, we first analyze the degree of discrepancy between the recovered data and the real data under the strict low-rank assumption, based on the rank-approximation error representation (see Theorem \ref{th.1}). 
Secondly, we deeply analyze the structural features of the error data, design a prior function to characterize the approximation error, and a new decomposition architecture (low-rank and overlapping group error representation) for real data is proposed. Finally, based on this new matrix data decomposition framework, a class of generalized low-rank matrix complementary models and corresponding solution algorithms are proposed.

\subsection{Overlapping Group Error Representation}\label{sec.3.1}

\begin{theorem}[$ k $-rank Approximation Error \cite{vershynin2018high}]\label{th.1}
	For a given matrix $ {\bf{M}} \in\mathbb{R}^{n_1\times n_2} $ satisfying the SVD, $ {{\bf{M}}_k} $ is the best approximation matrix of rank $ k $ of $ {\bf{M}} $ . Then $ \left\| {{\bf{M}} - {{\bf{M}}_k}} \right\|_F^2 $ can be represented by the singular values $ {\sigma _i} $ of $ {\bf{M}} $.
\end{theorem}

The proof of Theorem \ref{th.1} is provided in Appendix \ref{app.1}.

Theorem \ref{th.1} describes the error transformations relation between a matrix and its best approximation matrix of rank $ k $. 
It is evident that as $ k $ decreases, the rank approximation error increases. 
There is a contradiction between the lower rank approximation representation and the rank approximation error. 
Specifically, a lower-rank approximation benefits LRMC but leads to a larger approximation error. 
We find that the root cause of this phenomenon is that the matrix data is not strictly low-rank and the approximation error is not negligible.
Consequently, characterizing the matrix solely based on its low rank is unreasonable.

To better measure the inherent information of data, we propose a novel decomposition framework for real data. The decomposition model is as follows,
\begin{equation}\label{eq.a3}
{{\bf{Y}} = {\bf{X}} + {\cal E}},
\end{equation}
where $ {\bf{Y}} $ is the real matrix, $ {\bf{X}} $ and $ {\cal E} $ are the low-rank component and error component, respectively. 
The decomposition model divides the matrix data into a low-rank component and an error component, thereby enabling a more accurate measurement of the sparse properties inherent in the data.

\begin{figure*}[!htbp]
	\centering
	\includegraphics[width=6in]{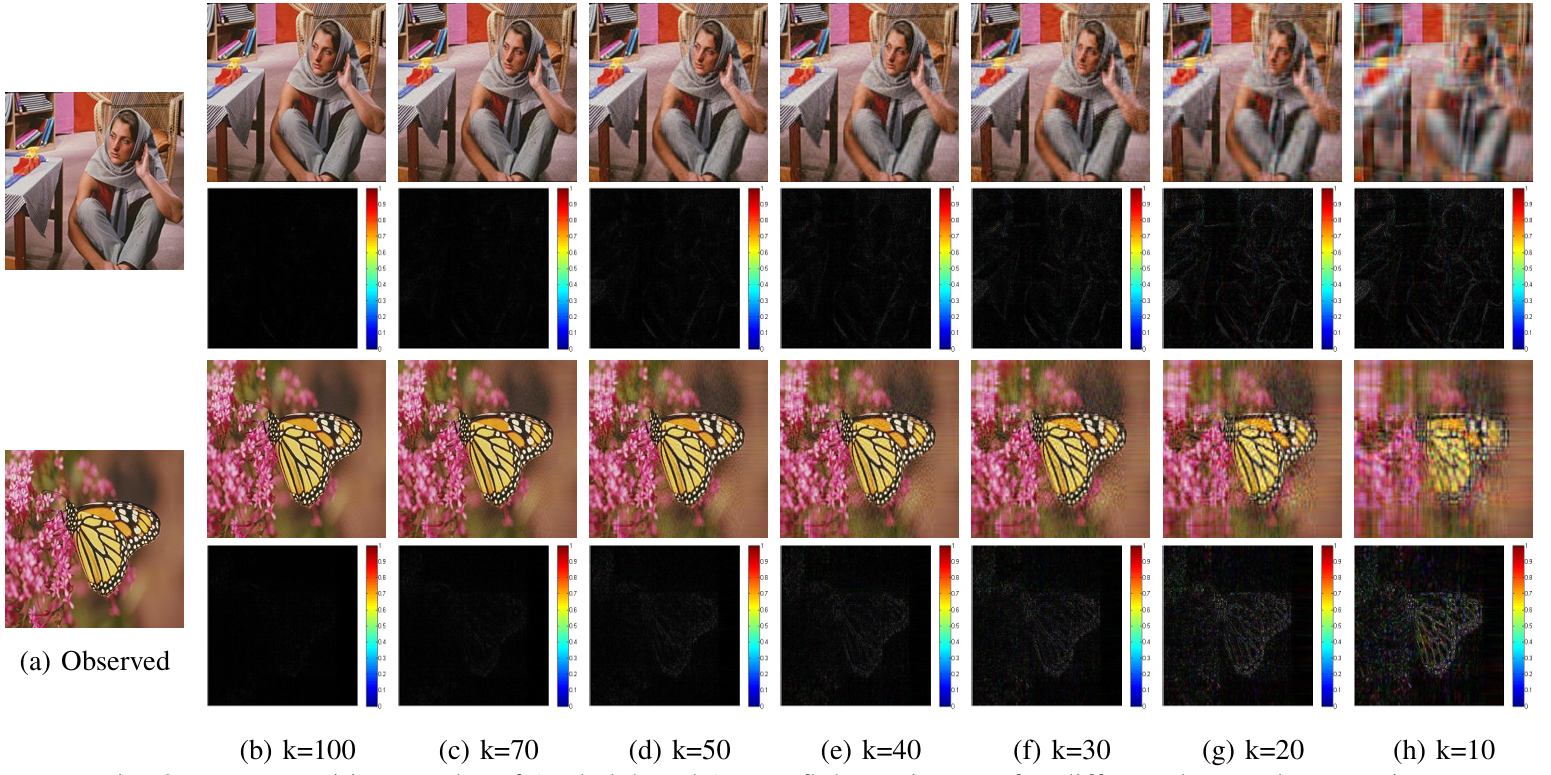}
\caption{Decomposition results of `Babala' and `Butterfly' test images for different low-rank truncations.}\label{fig.2}
\vspace{-0.2cm}
\end{figure*}

Therefore, to find a suitable priori function to describe the above error components, we have conducted some decomposition experiments. We selected two test images `Babala' and `Butterfly' and set the settings to different levels such as 100, 70, 50, 40, 30, 20, and 10. The corresponding low-rank and sparse components after interception are shown in Fig. \ref{fig.2}. Among them, the first and third rows are the low-rank components and the second and fourth rows are the sparse components. 
It can be observed from Fig. \ref{fig.2} that as the $ k $ value decreases, the approximation image tends to retain only information such as texture and details, making the overall effect of the image more blurred; Meanwhile, the error image tends to retain only information such as edges and contours, giving the image a more structured overall effect.

To address the significant structural block sparsity property of the error matrix data in the case of a smaller $ k $-value approximation, we design an overlapping group error representation (OGER) function to effectively characterize the structural block sparsity of the matrix.

\begin{definition}[OGER Function]\label{deoger}
	For any matrix  $ {\cal E}  \in\mathbb{R}^{n_1\times n_2} $, its OGER function $ \phi \left( {\cal E} \right) $ is defined as
\begin{equation}\label{eq.b1}
\begin{array}{l}
\vspace{0.05cm}
\phi \left( {\bf{\mathcal{E}}} \right) = \sum\nolimits_{i,j = 1} {{{\left[ {{{\sum\nolimits_{{k_1},{k_2} =  - {n_1}}^{{n_2}} {\left| {\varepsilon \left( {i + {k_1},j + {k_2}} \right)} \right|} }^2}} \right]}^{\frac{1}{2}}}} \\
\vspace{0.05cm}
\;\;\;\;\;\;\;\;\; = {\sum\nolimits_{i,j = 1} {\left\| {{{\bf{\mathcal{E} }}_{\left( {i,j} \right),K}}} \right\|} _2},
\end{array}
\end{equation}
where, $ {n_1} = \left[ {\frac{{K - 1}}{2}} \right] $, $ {n_2} = \left[ {\frac{K}{2}} \right] $, and $ {{\bf{\mathcal{E} }}_{\left( {i,j} \right),K}} $ is defined as a $ K*K $ point group at index set $ \left( {i,j} \right) $,
\[
\resizebox{\linewidth}{!}{$
	\begin{array}{l}
	{{\bf{\cal E }}_{\left( {i,j} \right),K}} = \\
	\left[ {\begin{array}{*{20}{c}}
		{\varepsilon \left( {i - {n_1},j - {n_1}} \right)}&{\varepsilon \left( {i - {n_1},j - {n_1} + 1} \right)}& \cdots &{\varepsilon \left( {i - {n_1},j + {n_2}} \right)}\\
		{\varepsilon \left( {i - {n_1} + 1,j - {n_1}} \right)}&{\varepsilon \left( {i - {n_1} + 1,j - {n_1} + 1} \right)}& \cdots &{\varepsilon \left( {i - {n_1} + 1,j + {n_2}} \right)}\\
		\vdots & \vdots & \ddots & \vdots \\
		{\varepsilon \left( {i + {n_2},j - {n_1}} \right)}&{\varepsilon \left( {i + {n_2},j - {n_1} + 1} \right)}& \cdots &{\varepsilon \left( {i + {n_2},j + {n_2}} \right)}
		\end{array}} \right].
	\end{array}
	$}
\]
\end{definition}

\begin{theorem}[OGER Convexity]\label{th.2}
	For any matrix  $ {\cal E}  \in\mathbb{R}^{n_1\times n_2} $, its overlapping group function $ \phi \left( {\cal E} \right) $ is a convex function.	
\end{theorem}
 
The proof of Theorem \ref{th.2} is provided in Appendix \ref{app.2}.

\begin{remark}\label{re.2}
OGER serves as an a priori function for inscribing matrices presenting sparse properties of structural blocks. Compared to the universal point-sparse ($ L^1 $-norm) representation, this inscription captures nonlocal sparse blocks in the error components more efficiently and describes the mathematical structure of such matrices more accurately. Furthermore, when $ K=1 $, the OGER function reduces to a universal point-sparse representation.
\end{remark}

\subsection{The Proposed Model and Algorithm}

In this subsection, we embed the OGER into the generalized low-rank matrix completion model and construct a novel generalized low-rank matrix completion model with overlapping group error representation. The proposed model utilizes the group sparsity metric to address the connection between pixel points in the local region, maintaining the local coherence and stability of the matrix, while effectively preserving the sparse texture and structure in the restored data to achieve superior restoration results. The specific model is as follows,
\begin{equation}\label{eq.b2}
\begin{array}{l}
\mathop {\min }\limits_{{\bf{X}},{\bf{\cal E }}} R\left( {\bf{X}} \right) + \lambda \phi \left( {\bf{\cal E }} \right)\\
s.t.\;{{\bf{Y}}_\Omega }{\rm{ = }}{{\bf{X}}_\Omega }{\rm{ + }}{{\bf{\cal E }}_\Omega },
\end{array}
\end{equation}
where $\lambda$ is a positive parameter, $ {\bf{X}} $ and $ {\bf{\cal E }} $ are the low-rank component and structural error component, respectively. And $ {{\bf{Y}}_\Omega }$ is the observed matrix data, $ \Omega  $ is the indexed set of uncorrupted points, $ R\left( {\bf{X}} \right) $ is the approximation function of the rank function, and $ \phi \left( {\bf{\cal E }} \right) $ denotes the OGER function used to characterize the structural error. By the primal-dual theory \cite{zhang2022generalized}, Eq. \eqref{eq.b2} can be rewritten as
\begin{equation}\label{eq.b3}
\mathop {\min }\limits_{{\bf{X}},{\bf{\cal E }}} \alpha R\left( {\bf{X}} \right) + \lambda \phi \left( {\bf{\cal E }} \right){\rm{ + }}\frac{1}{2}\left\| {{{\bf{Y}}_\Omega } - {{\bf{X}}_\Omega } - {{\bf{\cal E }}_\Omega }} \right\|_F^2,
\end{equation}
where $\alpha $ and $\lambda $ represent the regularization parameters. 

The ADMM algorithm is used to solve the above problems. To begin with, we introduce auxiliary variables ${\bf{W}}$, ${\bf{E}}$, and ${{\bf{F}}_\Omega }$. Then, we reformulate the above formula by rewriting it as the following,
\begin{equation}\label{eq.b4}
\begin{array}{l}
\mathop {\min }\limits_{{\bf{X}},{\bf{\cal E }},{\bf{W}},{\bf{E}},{{\bf{F}}_\Omega }} \alpha R\left( {\bf{W}} \right) + \lambda \phi \left( {\bf{E}} \right){\rm{ + }}\frac{1}{2}\left\| {{{\bf{F}}_\Omega }} \right\|_F^2\\
s.t.\;{\bf{W}} = {\bf{X}},\;{\bf{E}} = {\bf{\cal E }},\;{{\bf{F}}_\Omega } = {{\bf{Y}}_\Omega } - {{\bf{X}}_\Omega } - {{\bf{\cal E }}_\Omega }.
\end{array}
\end{equation}
Therefore, the augmented Lagrangian function of the problem \eqref{eq.b4} can be written as

\begin{equation}\label{eq.b5}
\begin{array}{l}
L\left( {{\bf{X}},\;{\bf{W}},\;{\bf{\cal E }},\;{{\bf{F}}_\Omega },\;{\bf{E}},\;{{\bm{\mu }}_1},\;{{\bm{\mu }}_2},\;{{\bm{\mu }}_3}} \right) = \\
\alpha R\left( {\bf{W}} \right) + \lambda \phi \left( {\bf{E}} \right) + \frac{1}{2}\left\| {{{\bf{F}}_\Omega }} \right\|_F^2\\
+ \frac{\rho }{2}\left\| {{\bf{W}} - {\bf{X}} + \frac{{{{\bm{\mu }}_1}}}{\rho }} \right\|_F^2 + \frac{\rho }{2}\left\| {{\bf{E}} - {\bf{\cal E }} + \frac{{{{\bm{\mu }}_2}}}{\rho }} \right\|_F^2\\
+ \frac{\rho }{2}\left\| {{{\bf{F}}_\Omega } - \left( {{{\bf{Y}}_\Omega } - {{\bf{X}}_\Omega } - {{\bf{\cal E }}_\Omega }} \right) + \frac{{{{\bm{\mu }}_3}}}{\rho }} \right\|_F^2,
\end{array}
\end{equation}
where, ${{{\bm{\mu }}_1}}$, ${{{\bm{\mu} }}_2}$ and ${{{\bm{\mu }}_3}}$ are Lagrange multipliers, and $\rho  > 0$ is the penalty parameter. All subproblems are described as
\begin{equation}\label{eq.b6}
\resizebox{0.91\linewidth}{!}{$
\left\{ \begin{array}{l}
{{\bf{X}}^{k + 1}} = \mathop {\arg \min }\limits_{\bf{X}} L\left( {{\bf{X}},\;{{\bf{W}}^k},\;{{{\bf{\cal E}}}^k},\;{\bf{F}}_\Omega ^k,\;{{\bf{E}}^k},{\bf{\bm{\mu }}}_1^k,\;{\bf{\bm{\mu }}}_2^k,\;{\bf{\bm{\mu }}}_3^k} \right)\\
{{\bf{W}}^{k + 1}} = \mathop {\arg \min }\limits_{\bf{W}} L\left( {{{\bf{X}}^{k + 1}},\;{\bf{W}},\;{{\bf{\cal E}}^k},\;{\bf{F}}_\Omega ^k,\;{{\bf{E}}^k},{\bf{\bm{\mu }}}_1^k,\;{\bf{\bm{\mu }}}_2^k,\;{\bf{\bm{\mu }}}_3^k} \right)\\
{{\cal E}^{k + 1}} = \mathop {\arg \min }\limits_{\cal E} L\left( {{{\bf{X}}^{k + 1}},\;{{\bf{W}}^{k + 1}},\;{\cal E},\;{\bf{F}}_\Omega ^{k},{{\bf{E}}^k},\;{\bf{\bm{\mu }}}_1^k,\;{\bf{\bm{\mu }}}_2^k,\;{\bf{\bm{\mu }}}_3^k} \right)\\
{\bf{F}}_\Omega ^{{k + 1}} = \mathop {\arg \min }\limits_{{{\bf{F}}_\Omega }} L\left( {{{\bf{X}}^{k + 1}},\;{{\bf{W}}^{k + 1}},\;{{\cal E}^{k + 1}},\;{{\bf{F}}_\Omega },\;{{\bf{E}}^k},{\bf{\bm{\mu }}}_1^k,\;{\bf{\bm{\mu }}}_2^k,\;{\bf{\bm{\mu }}}_3^k} \right)\\
{{\bf{E}}^{k + 1}} = \mathop {\arg \min }\limits_{\bf{E}} L\left( {{{\bf{X}}^{k + 1}},\;{{\bf{W}}^{k + 1}},\;{{\cal E}^{k + 1}},\;{\bf{F}}_\Omega ^{{k + 1}},\;{\bf{E}},{\bf{\bm{\mu }}}_1^k,\;{\bf{\bm{\mu }}}_2^k,\;{\bf{\bm{\mu }}}_3^k} \right)\\
{\bf{\bm{\mu }}}_1^{k + 1} = {\bf{\bm{\mu }}}_1^k + \rho \left( {{{\bf{W}}^{k + 1}} - {{\bf{X}}^{k + 1}}} \right)\\
{\bf{\bm{\mu }}}_2^{k + 1} = {\bf{\bm{\mu }}}_2^k + \rho \left( {{{\bf{E}}^{k + 1}} - {{\bf{\cal E }}^{k + 1}}} \right)\\
{\bf{\bm{\mu }}}_3^{k + 1} = {\bf{\bm{\mu }}}_3^k + \rho \left( {{{\bf{F}}_\Omega ^{k + 1}} - {{\bf{Y}}_\Omega } + {\bf{X}}_\Omega ^{k + 1} + {\bf{\cal E }}_\Omega ^{k + 1}} \right)
\end{array} \right.
	$}.
\end{equation}

\textbf{1) ${\bf{X}}$-subproblem}

The ${\bf{X}}$-subproblem in the Eq. \eqref{eq.b6} can be written as
\begin{equation}\label{eq.b7}
\begin{array}{l}
{{\bf{X}}^{k + 1}} = \mathop {\arg \min }\limits_{\bf{X}} \frac{\rho }{2}\left\| {{{\bf{W}}^k} - {\bf{X}} + \frac{{{\bf{\bm{\mu }}}_1^k}}{\rho }} \right\|_F^2\\
\;\;\;\;\;\;\;\;\;\; + \frac{\rho }{2}\left\| {{\bf{F}}_\Omega ^k - \left( {{{\bf{Y}}_\Omega } - {{\bf{X}}_\Omega } - {\bf{\cal E }}_\Omega ^k} \right) + \frac{{{\bf{\bm{\mu }}}_3^k}}{\rho }} \right\|_F^2.
\end{array}
\end{equation}
The Eq. \eqref{eq.b7} can be simplified as
\begin{equation}\label{eq.b8}
{{\bf{X}}^{k + 1}} = \mathop {\arg \min }\limits_{\bf{X}} \frac{\rho }{2}\left\| {{\bf{X}} - {\bf{N}}} \right\|_F^2 + \frac{\rho }{2}\left\| {{{\bf{X}}_\Omega } - {{\bf{K}}_\Omega }} \right\|_F^2,
\end{equation}
where ${\bf{N}} = {{\bf{W}}^k} + \frac{{{\bm{\bm{\mu} }}_1^k}}{\rho }$ and ${{\bf{K}}_\Omega } = {{\bf{Y}}_\Omega } - {\bf{F}}_\Omega ^k - {\bf{\cal E }}_\Omega ^k - \frac{{{\bm{\bm{\mu} }}_3^k}}{\rho }$.

\noindent
Solving the above formula yields,
\begin{equation}\label{eq.b9}
{{\bf{X}}^{k + 1}} = {{\bf{N}}_{{\Omega ^c}}} + \frac{{{{\bf{K}}_\Omega } + {{\bf{N}}_\Omega }}}{2}.
\end{equation}

\textbf{2) ${\bf{W}}$-subproblem}

The ${\bf{W}}$-subproblem in the Eq. \eqref{eq.b6}  can be written as
\begin{equation}\label{eq.b10}
{{\bf{W}}^{k + 1}} = \mathop {\arg \min }\limits_{\bf{W}} \alpha R\left( {\bf{W}} \right) + \frac{\rho }{2}\left\| {{\bf{W}} - {{\bf{X}}^{k + 1}} + \frac{{{\bm{\bm{\mu} }}_1^k}}{\rho }} \right\|_F^2,
\end{equation}
where $R\left( {\bf{W}} \right)$ is an approximate function of the rank function. Using the proximal operator, the minimization problem can be succinctly expressed as follows,
\begin{equation}\label{eq.b11}
{{\bf{W}}^{k + 1}} = \mathop {\arg \min }\limits_{\bf{W}} Prox{_{\alpha \rho R}}\left( {{{\bf{X}}^{k + 1}} - \frac{{{\bm{\bm{\mu} }}_1^k}}{\rho }} \right).
\end{equation}
Finally, based on the Von-Neuman trace inequality \cite{mirsky1975trace}, we obtain the relevant conclusion of the singular value contraction(see Theorem \ref{th.3}). And the closed-form solution of the above problem is easy to obtain.

\begin{lemma}[Von-Neuman Trace Inequality \cite{mirsky1975trace}]\label{le.1}
Given two matrices ${\bf{M=}}{{\bf{U}}_{\bf{M}}}{{\bf{S}}_{\bf{M}}}{\bf{V}}_{\bf{M}}^{{T}}$, $ {\bf{N}}={{\bf{Q}}_{\bf{N}}}{{\bf{\Sigma }}_{\bf{N}}}{\bf{R}}_{\bf{N}}^{{T}} $, where $ {{\bf{M}}, {\bf{N}}} \in\mathbb{R}^{n_1\times n_2} $ , $ {{\bf{S}}_{\bf{M}}} = diag\left( {{s_1}\left( {\bf{M}} \right),{s_2}\left( {\bf{M}} \right), \cdots ,{s_r}\left( {\bf{M}} \right)} \right) $, and $ {{\bf{\Sigma }}_{\bf{N}}} = diag\left( {{\sigma _1}\left( {\bf{N}} \right),{\sigma _2}\left( {\bf{N}} \right), \cdots ,{\sigma _r}\left( {\bf{N}} \right)} \right) $. Then they satisfy $ tr\left( {{{\bf{M}}^{{T}}}{\bf{N}}} \right) \le tr\left( {{{s}}{{\left({\bf{M}} \right)}^{{T}}}{\bf{\sigma }}\left( {\bf{N}} \right)} \right)$, where ${{s}}\left({\bf{M}} \right)$ and ${\bf{\sigma }}\left( {\bf{N}} \right)$ are the descending singular values of $\bf{M}$ and $\bf{N}$, respectively.
\end{lemma}

\begin{theorem}\label{th.3}
Consider a minimization problem of the following form,
\begin{equation}\label{eq.b12}
\mathop {\min }\limits_{\bf{W}} \lambda \sum\nolimits_{i = 1}^{\min \left\{ {{n_1},{n_2}} \right\}} {\psi \left( {{\sigma _i}} \right)}  + \frac{\rho }{2}\left\| {{\bf{W}} - {\bf{D}}} \right\|_F^2,
\end{equation}
where $\psi$ a smooth function, $ {{\bf{W}}, {\bf{D}}} \in\mathbb{R}^{n_1\times n_2} $, $\sigma _i$ and $ {s _i}$ is the $ i $-th singular value of $ \bf{W} $ and $ \bf{D} $, respectively.
Suppose that the singular value decomposition of $ \bf{D} $ is expressed as $ {\bf{D}} = {{\bf{Q}}_{\bf{D}}}{{\bf{S}}_{\bf{D}}}{\bf{R}}_{\bf{D}}^{T} $. Note that the solution of the above optimization problem (\ref{eq.b12}) is ${\bf{W}^*}={{\bf{U}}_{\bf{W}}}{{\bf{\Sigma}}_{\bf{W}}}{\bf{V}}_{\bf{W}}^{T}$, where $ {{\bf{U}}_{\bf{W}}} = {{\bf{Q}}_{\bf{D}}} $, $ {{\bf{V}}_{\bf{W}}} = {{\bf{R}}_{\bf{D}}} $, and  $ {{\bf{\Sigma }}_{\bf{W}}} $  is a diagonal matrix composed of ${\sigma}_i$, which can be solved by the following problem
\begin{equation}\label{eq.b13}
\mathop {\min }\limits_{{\sigma _i}} \lambda \psi \left( {{\sigma _i}} \right) + \frac{\rho }{2}{\left( {{\sigma _i} - {s_i}} \right)^2}.
\end{equation}
\end{theorem}

The proof of Theorem \ref{th.3} is provided in Appendix \ref{app.3}.

\begin{remark}\label{re.3}
In the literature \cite{candes2010power} and \cite{li2020matrix}, the optimal solutions of the two minimization problems are introduced in detail when ${R_1}\left( \bf{W} \right) = {\left\| \bf{W} \right\|_*}$ and ${R_2}\left( \bf{W} \right) = {\left\| \bf{W} \right\|_{\emph{Sp},\tau}}$, respectively.
\end{remark}

\textbf{3) $ {\bf{\cal E }} $-subproblem}

The $ {\bf{\cal E }} $-subproblem in the Eq. \eqref{eq.b6}  can be expressed as

\begin{equation}\label{eq.b14}
\begin{array}{l}
{{\bf{\cal E }}^{k + 1}} = \mathop {\arg \min }\limits_{\bf{\cal E }} \frac{\rho }{2}\left\| {{{\bf{E}}^k} - \bf{\cal E } + \frac{{{\bm{\mu }}_2^k}}{\rho }} \right\|_F^2\\
\;\;\;\;\;\;\;\;\; + \frac{\rho }{2}\left\| {{\bf{F}}_\Omega ^k - \left( {{{\bf{Y}}_\Omega } - {\bf{X}}_\Omega ^{k + 1} - {{\bf{\cal E }}_\Omega }} \right) + \frac{{{\bm{\mu }}_3^k}}{\rho }} \right\|_F^2.
\end{array}
\end{equation}

\noindent
The above formula can be simplified as
\begin{equation}\label{eq.b15}
{{\bf{\cal E }}^{k + 1}} = \mathop {\arg \min }\limits_{\bf{\cal E }} \frac{\rho }{2}\left\| {{\bf{\cal E }} - {\bf{G}}} \right\|_F^2 + \frac{\rho }{2}\left\| {{{\bf{\cal E }}_\Omega } - {{\bf{B}}_\Omega }} \right\|_F^2,
\end{equation}
where ${\bf{G}} = {{\bf{E}}^k} + \frac{{{\bm{\mu }}_2^k}}{\rho }$ and ${{\bf{B}}_\Omega } = {{\bf{Y}}_\Omega } - {\bf{X}}_\Omega ^{k + 1} - {\bf{F}}_\Omega ^k - \frac{{{\bm{\mu }}_3^k}}{\rho }$.

\noindent
Solve the above formula can be obtained
\begin{equation}\label{eq.b16}
{{\bf{\cal E }}^{k + 1}} = {{\bf{G}}_{{\Omega ^c}}} + \frac{{{{\bf{B}}_\Omega } + {{\bf{G}}_\Omega }}}{2}.
\end{equation}

\textbf{4) ${{{\bf{F}}_\Omega }}$-subproblem}

The ${{{\bf{F}}_\Omega }}$-subproblem in the Eq. \eqref{eq.b6} can be written as
\begin{equation}\label{eq.b17}
\begin{array}{l}
{{\bf{F}}_\Omega ^{k + 1}} = \mathop {\arg \min }\limits_{{{\bf{F}}_\Omega }} \frac{1}{2}\left\| {{{\bf{F}}_\Omega }} \right\|_F^2\\
\;\;\;\;\;\;\;\; + \frac{\rho }{2}\left\| {{{\bf{F}}_\Omega } - \left( {{{\bf{Y}}_\Omega } - {\bf{X}}_\Omega ^{k + 1} - {\bf{\cal E }}_\Omega ^{k + 1}} \right) + \frac{{{\bm{\mu }}_3^k}}{\rho }} \right\|_F^2.
\end{array}
\end{equation}
Solving the above formula yields
\begin{equation}\label{eq.b18}
{{\bf{F}}_\Omega ^{k + 1}} = \frac{{\rho \left( {{{\bf{Y}}_\Omega } - {\bf{X}}_\Omega ^{k + 1} - {\bf{\cal E }}_\Omega ^{k + 1}} \right) - {\bm{\mu }}_3^k}}{{1 + \rho }}.
\end{equation}

\textbf{5) ${\bf{E}}$-subproblem}

The ${\bf{E}}$-subproblem in the Eq. \eqref{eq.b6} can be written as
\begin{equation}\label{eq.b19}
{{\bf{E}}^{k + 1}} = \mathop {\arg \min }\limits_{\bf{E}} \lambda \phi \left( {\bf{E}} \right) + \frac{\rho }{2}\left\| {{\bf{E}} - {{\cal E }^{k + 1}} + \frac{{{{\bm{\mu }}_2^k}}}{\rho }} \right\|_F^2.
\end{equation}
Using MM algorithm, the optimal solution of the problem is
\begin{equation}\label{eq.b20}
{{\bf{E}}^{k + 1}} = {\left( {{\bf{I}} + \frac{{\lambda {\bf{\Lambda }}{{\left( {{{\bf{E}}^k}} \right)}^T}{\bf{\Lambda }}\left( {{{\bf{E}}^k}} \right)}}{\rho }} \right)^{ - 1}}{{\bf{D}}^k},
\end{equation}
where ${{\bf{D}}^k} = {{\bf{\cal E }}^{k + 1}} + \frac{{{\bm{\mu }}_2^k}}{\rho }$, ${\bf{\Lambda }}\left( {{{\bf{E}}^k}} \right)$ is a diagonal matrix, and the elements on its diagonal are expressed as
\[
\resizebox{\linewidth}{!}{$
{\left[ {{\bf{\Lambda }}\left( {{{\bf{E}}^k}} \right)} \right]_{l,l}} = \sqrt {{{\sum\limits_{i,j =  - {m_1}}^{{m_2}} {\left[ {\sum\limits_{{k_1},{k_2} =  - {m_1}}^{{m_2}} {{{\left| {{{\bf{E}}^k}_{r - i + {k_1},i - j + {k_2}}} \right|}^2}} } \right]} }^{ - 1/2}}}.
	$}
\]

The detailed solution process of the minimization problem is provided in Appendix \ref{app.a1}.

\textbf{6) Updating Lagrangian multipliers}

Finally, the Lagrangian multipliers are updated as follows,
\begin{equation}\label{eq.b21}
{\bm{\mu }}_1^{k + 1} = {\bm{\mu }}_1^k + \rho \left( {{{\bf{W}}^{k + 1}} - {{\bf{X}}^{k + 1}}} \right),
\end{equation}
\begin{equation}\label{eq.b22}
{\bm{\mu }}_2^{k + 1} = {\bm{\mu }}_2^k + \rho \left( {{{\bf{E}}^{k + 1}} - {{\bf{\cal E }}^{k + 1}}} \right),
\end{equation}
\begin{equation}\label{eq.b23}
{\bm{\mu }}_3^{k + 1} = {\bm{\mu }}_3^k + \rho \left( {{{\bf{F}}_\Omega ^{k + 1}} - {{\bf{Y}}_\Omega } + {\bf{X}}_\Omega ^{k + 1} + {\bf{\cal E }}_\Omega ^{k + 1}}\right).
\end{equation}

\begin{algorithm}[tbp]
	\setstretch{0.5}
	\renewcommand{\algorithmicrequire}{ \textbf{Input}:}
	\renewcommand{\algorithmicensure}{ \textbf{Output}:}
	\caption{Low-rank matrix completion with OGER (\ref{eq.b2})}\label{alg.1}
	\begin{algorithmic}[1]
		\REQUIRE $ \alpha $, $ \lambda $, $ {{\bf{Y}}_\Omega } $, $ N $, $K$ and $\rho  > 0$.
		\STATE Initialize $ {\bf{X}} $, $ {\bf{W}} $, $ {\bf{\cal E }} $, $ {{\bf{F}}_\Omega } $, $ {\bf{E}} $, $ {{\bm{\mu }}_1} $, $ {{\bm{\mu }}_2} $, $ {{\bm{\mu }}_3} $ and $ k = 0 $.
		\STATE \textbf{while} not converged \textbf{do}
		\STATE \quad Update ${{\bf{X}}^{k + 1}}$ by \eqref{eq.b9};
		\STATE \quad Update ${{\bf{W}}^{k + 1}}$ by \eqref{eq.b11};
		\STATE \quad Update ${{\bf{\cal E }}^{k + 1}}$ by \eqref{eq.b16};
		\STATE \quad Update ${{\bf{F}}_\Omega ^{k + 1}}$ by \eqref{eq.b18};
		\STATE \quad Update ${{\bf{E}}^{k + 1}}$ by \eqref{eq.b20};
		\STATE \quad Update ${\bm{\mu }}_1^{k + 1}$ , ${\bm{\mu }}_2^{k + 1}$  and ${\bm{\mu }}_3^{k + 1}$ by \eqref{eq.b21}, \eqref{eq.b22} and \eqref{eq.b23};
		\STATE \quad $ {{\bf{Y}}^{k + 1}} = {{\bf{X}}^{k + 1}} + {{\cal E}^{k + 1}} $;
		\STATE \quad $k=k+1$.
		\STATE \textbf{end while}
		\vspace{0.3em}
		\ENSURE ${{\bf{Y}}^*}={{\bf{Y}}^{k + 1}}$.
	\end{algorithmic}
\end{algorithm} 

\begin{remark}\label{re.4}
	The algorithm framework of the proposed model is shown in Algorithm \ref{alg.1}. The main cost of Algorithm \ref{alg.1} lies in computing $\bf{W}$ and $\bf{\cal E}$ subproblems when the ADMM framework is iterated once. In $\bf{W}$ subproblem, the complexity is mainly from computing the SVD and the shrinkage operator, whose computational costs are $O\left( {\min \left\{ {n_1^2{n_2},{n_1}n_2^2} \right\}} \right)$ and $O\left( {\min \left\{ {{n_1},{n_2}} \right\}} \right)$, respectively. In $\bf{\cal{E}}$ subproblem, the complexity is mainly from computing the  overlapping blocks of the matrix and the number of inner turns $N$, and their total computational cost is $O\left( {N{n_1}{n_2}} \right)$. Therefore, the computational cost at each iteration of Algorithm \ref{alg.1} is $O\left( {\min \left\{ {n_1^2{n_2},{n_1}n_2^2} \right\} + N{n_1}{n_2}} \right)$.
\end{remark}

\section{Numerical experiments}\label{sec.4}

In this section, to verify the effectiveness and superiority of the proposed low-rank matrix completion model with overlapping group error representation for image recovery, we choose two approximate representations of the rank function, $ R_1 $ \cite{candes2010power} and $ R_2 $ \cite{li2020matrix}. At this point, the model \eqref{eq.b2} is described in detail as

\[\mathop {\min }\limits_{{\bf{X}},{\bf{\cal E }}} {\left\| {\bf{X}} \right\|_*} + \lambda \phi \left( {\bf{\cal E }} \right)\;\;\;s.t.\;\;\;{\rm{ }}{{\bf{Y}}_\Omega }{\rm{ = }}{{\bf{X}}_\Omega }{\rm{ + }}{{\bf{\cal E }}_\Omega }.\]
\[\mathop {\min }\limits_{{\bf{X}},{\bf{\cal E }}} {\left\| {\bf{X}} \right\|_{Sp,\tau }} + \lambda \phi \left( {\bf{\cal E }} \right)\;\;\;s.t.\;\;\;{\rm{ }}{{\bf{Y}}_\Omega }{\rm{ = }}{{\bf{X}}_\Omega }{\rm{ + }}{{\bf{\cal E }}_\Omega }.\]

We compare the two new models with four other classical models, including the nuclear norm minimization (NNM) model \cite{candes2010power}, the weighted schatten $ p $ norm minimization (WSP) model \cite{xie2016weighted}, the schatten capped $ p $ (SCP) method \cite{li2020matrix}, and the ratio of the nuclear norm and the frobenius norm (N/F) method \cite{gao2024low}. We conducted a series of matrix completion experiments under identical conditions to assess the image recovery capabilities of these six models. 
All experiments were performed on a computer equipped with a Windows 10 operating system and an Intel Core i5-8250U running at 1.80 GHz and completed using MATLAB (R2014a version) software.

\begin{figure}[!t]
	\centering
	\includegraphics[width=3in]{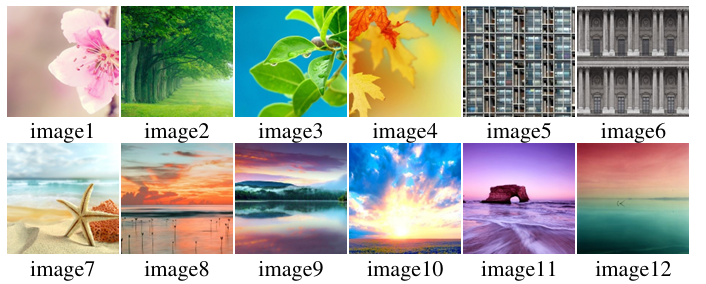}
	\caption{Test images for numerical experiments.}\label{fig.3}
	\vspace{-0.5cm}
\end{figure}

Fig. \ref{fig.3} presents 12 test images, which have been widely used for assessing the effects of image recovery across various models due to their low-rank nature. To evaluate the recovery effect of the models more effectively, two objective evaluation metrics, peak signal-to-noise ratio (PSNR) and signal-to-noise ratio (SNR) are employed for the experimental results. The definitions of PSNR and SNR are as follows,
\[{\rm{PSNR}} = 10 \cdot {\log _{10}}\left( {\frac{{{n_1}{n_2}}}{{\left\| {{\bf{Y}} - {{\bf{Y}}^ * }} \right\|_F^2}}} \right),\]

\[{\rm{SNR}} = 10 \cdot {\log _{10}}\left( {\frac{{\left\| {\bf{Y}} \right\|_F^2}}{{\left\| {{\bf{Y}} - {{\bf{Y}}^ * }} \right\|_F^2}}} \right).\]

In the numerical experiments, the higher the PSNR and SNR values relative to each other, the better the recovered image. The bold font indicates the optimal value. 
The ratio between the observed matrix elements and the total number of elements in the entire matrix is expressed using the sampling rate $ \eta  $, which is defined as
\[\eta  = 1 - \frac{{num\left( {{{\bf{Y}}_\Omega }} \right)}}{{num\left( {\bf{Y}} \right)}},\]
where $num\left(  \cdot  \right) $ denotes the number of entries in the matrix.
Furthermore, the iterative stopping condition of the modeling algorithm is set to ensure that the relative error (RE) of the recovered image in two consecutive iterations is less than the preset accuracy.
Here, the preset accuracy is set to $ tol = {10^{ - 5}} $,
\[RE = \frac{{\left\| {{{\bf{Y}}^k} - {{\bf{Y}}^{k - 1}}} \right\|_F^2}}{{\left\| {{{\bf{Y}}^k}} \right\|_F^2}} \le tol.\]

\subsection{Discussion of Related Parameters}

In this subsection, we delve into the parameters of the model in detail. Note that, we choose the surrogate function $ R_2 $ to represent the low-rank component within the model. 
The parameters under discussion primarily encompass the penalty parameter $\rho$, as well as the iteration number $ N $ and the group size $ K $  in the overlapping group subproblem \eqref{eq.b19}.

\begin{figure}[!t]
	\centering
	\includegraphics[width=3in]{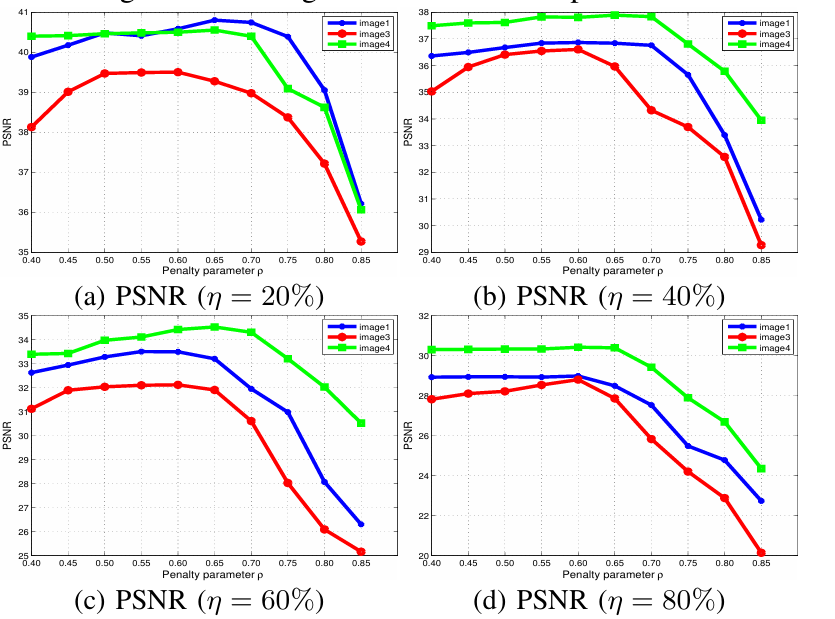}
	\caption{The PSNR values of the test images (`image1', `image3' and, `image4') are restored by different penalty parameter $\rho$ values.}\label{fig.4}
	\vspace{-0.2cm}
\end{figure}

1) \textit{Penalty parameter} $\rho$: We selected `image1', `image3', and `image4' as test images and applied random loss masks with values of $\eta  = 20\%, 40\%, 60\% $, $80\% $.
The parameter $\rho$ was varied within the range of $\left[ {0.4,0.85} \right]$ in increments of 0.05, while the remaining parameters remained fixed.
Fig. \ref{fig.4} illustrates the variation of the PSNR values across different values of $\rho$. We observed that the highest PSNR was achieved when $\rho=0.6$. Therefore, we set $\rho=0.6$ for the subsequent experiments.

2) \textit{Iteration number} $ N $: We select the test images including `image4'-`image9', to which a text mask 2 has been applied. 
We set $ N$ = $1$, $3$, $5$, $10$, $30$, $50 $ while keeping other parameters constant.
The resulting PSNR and SNR values for the recovered images are presented in Table \ref{table.1}.
From Table \ref{table.1}, it can be observed that as $ N $ increases from 1 to 5, the PSNR and SNR values gradually increase. However, once $ N $ exceeds 5, the PSNR and SNR values remain largely unchanged. Consequently, to avoid the additional time expenditure associated with excessive iterations, we set $ N = 5 $ for the subsequent experiments.

\begin{table*}[tp]
	\renewcommand{\arraystretch}{1}
	\setlength\tabcolsep{3.0pt}
	\footnotesize
	\caption{The PSNR and SSIM values for different $N$ values with text mask 2.}\label{table.1}
	\setlength{\abovecaptionskip}{5pt}
	\setlength{\belowcaptionskip}{5pt}
	\centering
	\vspace{-0.3cm}
	\begin{center}
		\resizebox{0.95\textwidth}{!}{
			\begin{tabular}{l||c|c|c|c|c|c|c|c|c|c|c|c}
				\Xhline{1pt}
				$N$ & \multicolumn{2}{c|}{$N=1$} & \multicolumn{2}{c|}{$N=3$} & \multicolumn{2}{c|}{$N=5$} & \multicolumn{2}{c|}{$N=10$} & \multicolumn{2}{c|}{$N=30$} &\multicolumn{2}{c}{$N=50$} \\
				\cline{2-13}
				Image & PSNR & SNR & PSNR & SNR & PSNR & SNR & PSNR & SNR & PSNR & SNR  & PSNR & SNR\\
				\Xhline{1pt}
				image4  &39.0170 & 29.6185 & 39.1890 & 29.8357 & 39.3219 & 30.1686 & 39.3119 & 30.1597 & 39.3075 & 30.1492 & 39.3008 & 30.1416  \\
				image5  &37.0144 & 31.4921 & 37.1419 & 31.6133 & 37.3028 & 31.7674 & 37.3101 & 31.7732 & 37.3943 & 31.7580 & 37.2722 & 31.7365  \\
				image6  &40.1568 & 33.7584 & 40.3102 & 33.9193 & 40.4613 & 34.0681 & 40.4698 & 34.1057 & 40.4574 & 34.0926 & 40.4229 & 34.0772 \\
				image7  &40.0822 & 31.5669 & 40.1731 & 31.7562 & 40.3174 & 37.8480 & 40.3029 & 37.8296 & 40.2881 & 37.8012 & 40.2763 & 37.7949 \\
				image8  &40.3058 & 34.5012 & 40.4918 & 34.6716 & 40.6893 & 34.8816 & 40.6914 & 34.8922 & 40.6830 & 34.8675 & 40.6715 & 34.8534 \\
				image9  &39.3870 & 34.5911 & 39.6375 & 34.7445 & 39.7685 & 34.9089 & 39.7613 & 34.8971 & 39.7524 & 34.8783 & 39.7316 & 34.8608 \\ 	
				\Xhline{1pt}
			\end{tabular}
		}
	\end{center}
	\vspace{-0.2cm}
\end{table*}

\begin{figure}[!t]
	\centering
	\includegraphics[width=3in]{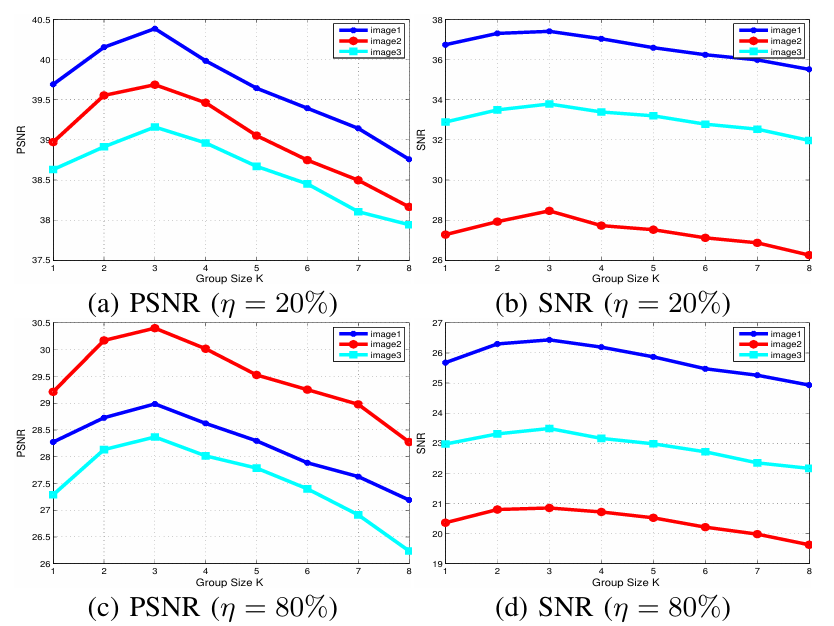}
	\caption{The PSNR and SNR values of the test images (`image1', `image2', and `image3') are restored by different group size $ K $ values.}\label{fig.5}
	\vspace{-0.2cm}
\end{figure}

3) \textit{Group size} $K$: We selected `image1', `image2', and `image3' as test images and applied a random loss mask with values of $\eta  = 20\% $ and $\eta  = 80\% $.
The parameter $ K $ was varied within the range of $\left[ {1,8} \right]$ in increments of 1, while the remaining parameters remained fixed. 
Fig. \ref{fig.5} illustrates the functional relationship between the PSNR and SNR values of the restored images and various values of $ K $.
It is important to note that when $ K = 1 $, the overlapping group sparsity metric reduces to a point sparsity metric.
We observed that the proposed model achieved the best numerical results when $ K = 3 $. Therefore, we set $ K = 3 $ for subsequent experiments.

Furthermore, other parameters within the model were manually adjusted to achieve a more optimal recovery outcome. The regularization parameters were set to $\alpha  \in \left[ {0.1,5} \right]$ and $\lambda  \in \left[ {0.01,1} \right]$.

\subsection{Random Mask Experiments}
\begin{figure}[!t]
	\centering
	\includegraphics[width=3in]{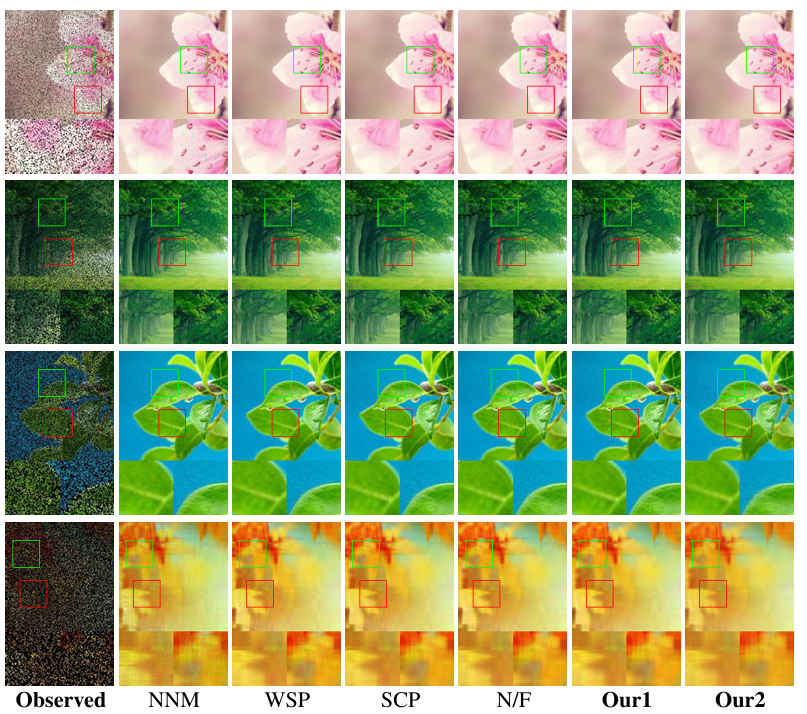}
	\caption{The restored results of different models for random mask experiments.}\label{fig.6}
	\vspace{-0.5cm}
\end{figure}

In this subsection, the recovery effect of the proposed model is validated with the addition of random masks. We apply masks with randomly missing entries of $ \eta$ = $ 20\% $, $ 40\% $, $ 60\% $, $ 80\% $ to the test images and present the recovered PSNR and SNR values of the six models in Table \ref{table.2}.
The numerical results in Table \ref{table.2} indicate that the proposed model outperforms the other models. Due to space limitations, Fig. \ref{fig.6} shows only a selection of the recovered images. It can be observed that as the number of missing entries in the image increases, the recovery effect of the model decreases.
When the missing entries reached $80\%$, the SCP model and the N/F model left a small amount of mask noise in the recovered image, in contrast to the proposed model. Consequently, the numerical results and subjective visualization both demonstrate that the proposed model outperforms the other competing models in terms of recovery.

\begin{table*}[tp]
	\renewcommand{\arraystretch}{1}
	\setlength\tabcolsep{3.0pt}
	\footnotesize
	\caption{The PSNR and SNR values of all methods with different random masks.}\label{table.2}
	\setlength{\abovecaptionskip}{5pt}
	\setlength{\belowcaptionskip}{5pt}
	\centering
	\vspace{-0.3cm}
		\begin{center}
		\resizebox{0.95\textwidth}{!}{
	\begin{tabular}{l||c|c|c|c|c|c|c|c|c|c|c|c|c} 
		\Xhline{1pt}
		Methods &  \multirow{2}{*}{mask}& \multicolumn{2}{c|}{NNM} & \multicolumn{2}{c|}{WSP} & \multicolumn{2}{c|}{SCP} & \multicolumn{2}{c|}{N/F} & 
		\multicolumn{2}{c|}{Our1} &\multicolumn{2}{c}{Our2} \\
		\cline{3-14}
		Image & & PSNR & SNR & PSNR & SNR & PSNR & SNR & PSNR & SNR & PSNR & SNR  & PSNR & SNR\\
			\Xhline{1pt}			
\multirow{4}{*}{image1}  & random mask 20\% & 36.18  & 32.92  & 38.40  & 35.29  & 39.77  & 36.62  & 40.26  & 37.24  & 38.69  & 35.54  & \textbf{40.38}  & \textbf{37.41}\\
                         &random mask 40\%  & 32.58  & 29.31  & 34.14  & 31.75  & 35.04  & 32.80  & 35.93  & 33.12  & 34.58  & 32.05  & \textbf{36.62}  & \textbf{33.87}\\
                         &random mask 60\%  & 29.12  & 26.56  & 31.27  & 27.98  & 32.25  & 29.16  & 32.65  & 29.42  & 31.65  & 28.18  & \textbf{33.04}  & \textbf{30.06}\\
                         &random mask 80\%  & 24.57  & 22.19  & 26.30  & 24.92  & 27.97  & 25.64  & 28.42  & 26.08  & 26.89  & 25.27  & \textbf{28.98}  & \textbf{26.41}\\
                        \hline
\multirow{4}{*}{image2}  &random mask 20\% & 35.50  & 24.75  & 37.74  & 26.35  & 38.90  & 27.98  & 39.45  & 28.20  & 37.81  & 26.42  & \textbf{39.69}  & \textbf{28.63}\\
                         &random mask 40\% & 32.98  & 22.06  & 34.01  & 23.52  & 35.56  & 24.28  & 35.91  & 24.93  & 34.44  & 23.62  & \textbf{36.48}  & \textbf{25.42}\\
                         &random mask 60\% & 29.87  & 20.10  & 31.39  & 21.57  & 32.43  & 22.32  & 32.86  & 22.79  & 31.58  & 21.43  & \textbf{33.32}  & \textbf{23.14}\\
                         &random mask 80\% & 26.17  & 17.29  & 28.62  & 19.16  & 29.87  & 20.13  & 30.04  & 20.22  & 28.97  & 19.65  & \textbf{30.40}  & \textbf{20.89}\\
                         \hline
\multirow{4}{*}{image3}  &random mask 20\% & 35.21  & 30.85  & 37.24  & 32.02  & 38.43  & 32.95  & 38.87  & 33.12  & 37.60  & 32.57  & \textbf{39.15}  & \textbf{33.78} \\
                         &random mask 40\% & 32.68  & 26.05  & 33.99  & 28.61  & 35.08  & 29.72  & 35.22  & 29.85  & 34.13  & 28.95  & \textbf{36.62}  & \textbf{30.54}\\
                         &random mask 60\% & 29.09  & 24.13  & 30.60  & 24.86  & 31.15  & 25.58  & 31.43  & 25.96  & 30.92  & 25.38  & \textbf{32.04}  & \textbf{26.91}\\
                         &random mask 80\% & 25.46  & 19.09  & 27.45  & 21.91  & 27.96  & 22.87  & 28.12  & 23.06  & 27.83  & 22.16  & \textbf{28.67}  & \textbf{23.48}\\
                       \hline
\multirow{4}{*}{image4}  &random mask 20\% & 36.69  & 27.25  & 38.76  & 29.33  & 39.40  & 30.05  & 39.64  & 30.26  & 39.03  & 29.60  & \textbf{40.06}  & \textbf{30.91}\\
                         &random mask 40\% & 34.00  & 24.09  & 35.56  & 26.48  & 36.82  & 27.53  & 37.06  & 28.02  & 35.74  & 26.59  & \textbf{37.67}  & \textbf{28.50}\\
                         &random mask 60\% & 30.81  & 22.13  & 32.25  & 23.49  & 33.97  & 24.89  & 34.15  & 25.03  & 32.39  & 23.73  & \textbf{34.74}  & \textbf{25.58}\\
                         &random mask 80\% & 26.23  & 18.04  & 27.98  & 19.96  & 29.43  & 21.62  & 29.43  & 21.74  & 28.15  & 20.21  & \textbf{30.25}  & \textbf{22.06}\\
		\Xhline{1pt}
	\end{tabular}}
\end{center}
	\vspace{-0.2cm}
\end{table*}

\subsection{Text Mask Experiments}

\begin{figure}[!t]
	\centering
	\includegraphics[width=3in]{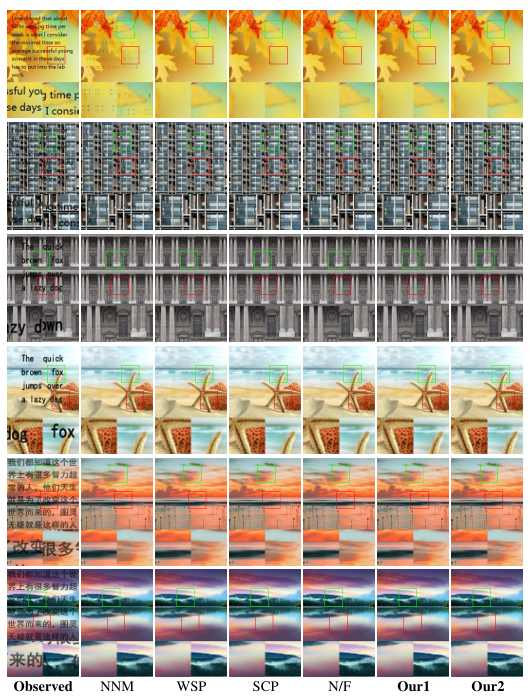}
	\caption{The restored results of different models for text mask experiments.}\label{fig.7}
	\vspace{-0.5cm}
\end{figure}

To further verify the effectiveness of the proposed model in recovering important information, such as details and textures, we recover images with added text masks in this subsection.
The numerical results of the recovered images for the six models are shown in Table \ref{table.3}. The results indicate that the proposed model is capable of achieving optimal recovery outcomes compared to other competing models. Fig. \ref{fig.7} shows only a selection of the recovered images from Table \ref{table.3}.
The numerical results and the quality of the recovered images using the proposed model 1 are demonstrably superior to those of the NNM model. Some of the images are recovered to a higher standard than those of the WSP model. This suggests that the proposed model, which incorporates overlapping group error characterization, exhibits superior performance.

\begin{table*}[tp]
	\renewcommand{\arraystretch}{1}
	\setlength\tabcolsep{3.0pt}
	\footnotesize
	\caption{ The PSNR and SNR values of all methods with different text masks.}\label{table.3}
	\setlength{\abovecaptionskip}{5pt}
	\setlength{\belowcaptionskip}{5pt}
	\centering
	\vspace{-0.3cm}
		\begin{center}
		\resizebox{0.95\textwidth}{!}{
	\begin{tabular}{l||c|c|c|c|c|c|c|c|c|c|c|c|c} 
		\Xhline{1pt}
		Methods &  \multirow{2}{*}{mask}& \multicolumn{2}{c|}{NNM} & \multicolumn{2}{c|}{WSP} & \multicolumn{2}{c|}{SCP} & \multicolumn{2}{c|}{N/F} & 
		\multicolumn{2}{c|}{Our1} &\multicolumn{2}{c}{Our2} \\
		\cline{3-14}
		Image & & PSNR & SNR & PSNR & SNR & PSNR & SNR & PSNR & SNR & PSNR & SNR  & PSNR & SNR\\
		\Xhline{1pt}

\multirow{4}{*}{image4} & text mask1 & 35.14  & 26.81  & 37.77  & 28.61  & 38.68  & 29.42  & 39.09  & 29.66  & 38.02  & 28.94  & \textbf{39.32}  & \textbf{30.16} \\
		                & text mask2 & 35.36  & 27.29  & 37.82  & 28.76  & 38.84  & 29.93  & 39.22  & 30.03  & 38.33  & 29.12  & \textbf{39.51}  & \textbf{30.24}\\
		                & text mask3 & 35.22  & 26.93  & 37.80  & 28.72  & 38.79  & 29.78  & 39.15  & 29.82  & 38.19  & 28.99  & \textbf{39.35}  & \textbf{30.20} \\
		  \hline
\multirow{4}{*}{image5}	& text mask1 & 33.14  & 28.06  & 35.39  & 29.86  & 36.93  & 31.27  & 37.01  & 31.43  & 35.54  & 30.05  & \textbf{37.29}  & \textbf{31.76}\\
		                & text mask2 & 35.42  & 29.89  & 37.21  & 31.68  & 38.16  & 32.63  & 38.32  & 32.79  & 37.52  & 31.99  & \textbf{38.54}  & \textbf{33.02}\\
		                & text mask3 & 33.82  & 28.31  & 35.94  & 30.41  & 37.72  & 32.19  & 37.96  & 32.43  & 36.09  & 30.56  & \textbf{38.13}  & \textbf{32.60}\\
		  \hline
\multirow{4}{*}{image6}	& text mask1 & 36.34  & 29.85  & 38.47  & 32.02  & 39.66  & 33.28  & 40.14  & 33.56  & 38.55  & 32.10  & \textbf{40.47}  & \textbf{34.02}\\
		                & text mask2 & 37.23  & 30.78  & 39.58  & 33.04  & 40.52  & 34.08  & 41.07  & 34.62  & 39.63  & 33.49  & \textbf{41.45}  & \textbf{35.01} \\
		                & text mask3 & 37.08  & 30.63  & 39.34  & 32.89  & 40.34  & 33.95  & 40.82  & 34.37  & 39.86  & 33.42  & \textbf{41.23}  & \textbf{34.78}  \\
\hline
\multirow{4}{*}{image7}	& text mask1 & 36.03  & 33.56  & 38.85  & 36.38  & 39.46  & 36.98  & 39.89  & 37.42  & 39.11  & 36.64  & \textbf{40.32}  & \textbf{37.85} \\
		                & text mask2 & 37.36  & 34.92  & 39.77  & 37.10  & 40.43  & 37.96  & 40.73  & 38.27  & 39.90  & 37.41  & \textbf{41.39}  & \textbf{38.93} \\
                        & text mask3 & 37.35  & 34.88  & 39.45  & 36.99  & 40.26  & 37.79  & 40.60  & 38.13  & 39.68  & 37.23  & \textbf{41.18}  & \textbf{38.71}  \\
		  \hline
\multirow{4}{*}{image8} & text mask1 & 37.23  & 31.44  & 39.02  & 32.23  & 40.04  & 34.25  & 40.21  & 34.42  & 39.32  & 33.53  & \textbf{40.67}  & \textbf{34.88} \\
		                & text mask2 & 37.64  & 31.85  & 39.93  & 34.14  & 41.22  & 35.42  & 41.56  & 35.77  & 40.51  & 34.71  & \textbf{41.87}  & \textbf{36.08}  \\
		                & text mask3 & 37.33  & 31.54  & 39.51  & 33.72  & 40.99  & 35.21  & 41.36  & 35.57  & 40.12  & 34.33  & \textbf{41.62}  & \textbf{35.83} \\
		  \hline
\multirow{4}{*}{image9} & text mask1 & 36.20  & 31.33  & 38.65  & 33.79  & 39.17  & 34.31  & 39.41  & 34.54  & 39.98  & 34.11  & \textbf{39.77}  & \textbf{34.91} \\
		                & text mask2 & 36.83  & 31.97  & 38.88  & 34.01  & 39.89  & 35.03  & 40.27  & 35.36  & 39.35  & 34.48  & \textbf{40.43}  & \textbf{35.57}\\
		                & text mask3 & 36.60  & 31.62  & 38.71  & 33.85  & 39.31  & 34.45  & 39.71  & 34.85  & 39.05  & 34.30  & \textbf{40.00}  & \textbf{35.14}\\	
		\Xhline{1pt}
	\end{tabular}}
\end{center}
	\vspace{-0.2cm}
\end{table*}

\subsection{Block Mask Experiments}

\begin{figure}[!t]
	\centering
	\includegraphics[width=3in]{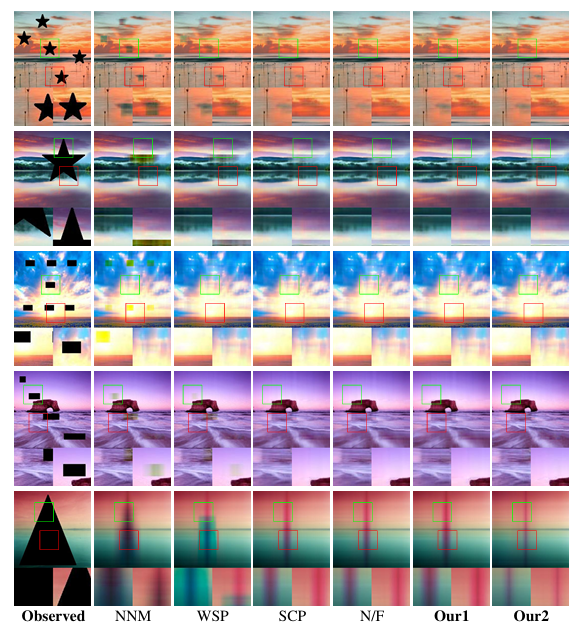}
	\caption{The restored results of different models for block mask experiments.}\label{fig.8}
	\vspace{-0.5cm}
\end{figure}

This subsection aims to corroborate the efficacy of the proposed model in addressing the challenges posed by extensive missing data in images. To achieve this, we conduct recovery experiments on images with added block masks.
Table \ref{table.4} illustrates the recovery of PSNR and SNR values for six models, with corresponding results presented in Fig. \ref{fig.8}.

\begin{figure}[!t]
	\centering
	\includegraphics[width=3in]{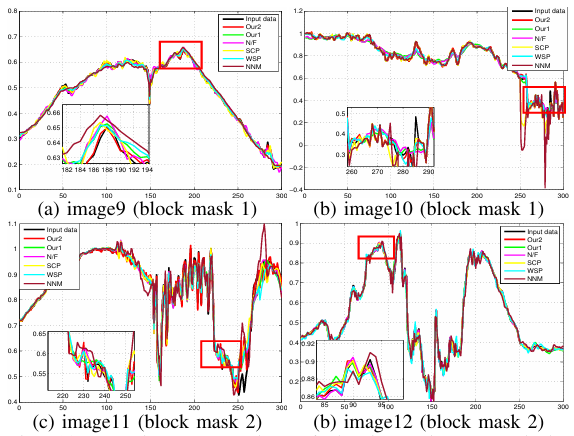}
	\caption{Local cross-ssection of the input image and the restored image.}\label{fig.9}
	\vspace{-0.2cm}
\end{figure}

The significant data loss in the image resulting from the application of the block mask poses a substantial challenge to the models' ability to recover the original information. 
For instance, when the triangular block mask is incorporated, the NNM model and the WSP model result in the formation of substantial black blocks in the central region of the recovered image, while the SCP model and the N/F model are unable to fully recover the intricate details of the tree branches when the large pentagram mask is employed. 
In contrast, the recovered images from the proposed model appear more naturalistic.
Furthermore, Fig. \ref{fig.9} illustrates localized slices of `image9'-`image12' in their $ 200 $th column after recovery. It can be seen that the proposed model produces the most effective results.

\begin{table*}[tp]
	\renewcommand{\arraystretch}{1}
	\setlength\tabcolsep{3.0pt}
	\footnotesize
	\caption{The PSNR and SNR values of all methods with different block masks.}\label{table.4}
	\setlength{\abovecaptionskip}{5pt}
	\setlength{\belowcaptionskip}{5pt}
	\centering
	\vspace{-0.2cm}
		\begin{center}
		\resizebox{0.95\textwidth}{!}{
	\begin{tabular}{l||c|c|c|c|c|c|c|c|c|c|c|c|c} 
		\Xhline{1pt}
		Methods &  \multirow{2}{*}{mask}& \multicolumn{2}{c|}{NNM} & \multicolumn{2}{c|}{WSP} & \multicolumn{2}{c|}{SCP} & \multicolumn{2}{c|}{N/F} & 
		\multicolumn{2}{c|}{Our1} &\multicolumn{2}{c}{Our2} \\
		\cline{3-14}
		Image & & PSNR & SNR & PSNR & SNR & PSNR & SNR & PSNR & SNR & PSNR & SNR  & PSNR & SNR\\
		\Xhline{1pt}
\multirow{4}{*}{image8} & block mask1 &		32.37  & 24.58  & 34.26  & 28.47  & 35.01  & 29.22  & 35.29  & 29.50  & 34.68  & 28.89  & \textbf{35.63}  & \textbf{29.85}  \\ 
& block mask2 & 		25.09  & 19.31  & 27.15  & 21.36  & 28.40  & 22.61  & 28.86  & 23.03  & 27.77  & 21.98  & \textbf{29.04}  & \textbf{23.21}  \\ 
& block mask3 & 		28.69  & 22.90  & 30.93  & 25.08  & 31.76  & 25.91  & 31.97  & 26.19  & 31.37  & 25.58  & \textbf{32.35}  & \textbf{26.56}  \\ 
& block mask4 & 		29.95  & 24.13  & 32.15  & 26.34  & 32.52  & 26.76  & 32.81  & 27.02  & 31.85  & 26.02  & \textbf{33.01}  & \textbf{27.24 } \\ 
& block mask5 & 		20.01  & 14.28  & 22.29  & 16.33  & 23.17  & 17.32  & 23.64  & 17.87  & 22.66  & 16.86  & \textbf{23.87}  & \textbf{18.05}  \\ 
 \hline
\multirow{4}{*}{image9} & block mask1 & 		31.79  & 26.61  & 33.95  & 28.89  & 34.85  & 29.99  & 35.06  & 30.19  & 34.41  & 29.54  & \textbf{35.42}  & \textbf{30.25}  \\
& block mask2 & 		24.60  & 19.74  & 26.94  & 22.07  & 27.96  & 23.10  & 28.19  & 23.33  & 27.32  & 22.46  & \textbf{28.44}  & \textbf{23.57}  \\ 
& block mask3 & 		28.29  & 23.12  & 30.64  & 25.41  & 31.54  & 26.37  & 32.08  & 26.91  & 30.95  & 25.78  & \textbf{32.42}  & \textbf{27.23}  \\ 
& block mask4 & 		30.00  & 25.13  & 32.62  & 27.45  & 33.42  & 28.26  & 33.85  & 28.67  & 33.21  & 28.04  & \textbf{34.02}  & \textbf{28.84}  \\ 
& block mask5 & 		17.53  & 12.57  & 19.95  & 14.88  & 21.06  & 15.89  & 21.54  & 16.37  & 20.54  & 15.33  & \textbf{21.88}  & \textbf{16.71}  \\ 
 \hline
\multirow{4}{*}{image10} & block mask1 & 		28.31  & 26.39  & 30.07  & 28.16  & 34.62  & 32.71  & 35.00  & 33.09  & 34.32  & 32.40  & \textbf{35.34}  & \textbf{33.41}  \\ 
& block mask2 & 		24.94  & 19.28  & 27.36  & 22.41  & 28.09  & 23.18  & 28.46  & 23.55  & 27.89  & 22.92  & \textbf{28.87}  & \textbf{22.96}  \\ 
& block mask3 & 		25.69  & 23.75  & 28.97  & 27.02  & 29.65  & 27.74  & 29.88  & 27.97  & 29.31  & 27.43  & \textbf{30.13}  & \textbf{27.25}  \\ 
& block mask4 & 		28.13  & 26.22  & 30.50  & 28.59  & 31.45  & 29.54  & 31.77  & 29.85  & 31.04  & 29.12  & \textbf{32.00}  & \textbf{30.88}  \\ 
& block mask5 & 		14.94  & 11.02  & 17.35  & 15.01  & 19.13  & 17.26  & 19.59  & 17.68  & 18.85  & 16.91  & \textbf{19.76}  & \textbf{17.85}  \\ 
 \hline
\multirow{4}{*}{image11} & block mask1 & 		27.12  & 25.10  & 30.43  & 28.42  & 31.01  & 28.97  & 31.35  & 29.33  & 30.70  & 28.69  & \textbf{31.56}  & \textbf{29.54 } \\ 
& block mask2 & 		21.94  & 17.86  & 24.05  & 20.98  & 24.87  & 21.88  & 25.12  & 22.10  & 24.32  & 21.29  & \textbf{25.45}  & \textbf{22.44}  \\ 
& block mask3 & 		25.03  & 23.09  & 27.87  & 25.80  & 28.48  & 26.46  & 28.80  & 26.78  & 28.04  & 26.02  & \textbf{29.13}  & \textbf{27.17}  \\ 
& block mask4 & 		26.76  & 24.75  & 29.51  & 27.49  & 30.24  & 28.21  & 30.65  & 28.62  & 29.92  & 27.90  & \textbf{30.89}  & \textbf{28.93}  \\ 
& block mask5 & 		13.19  & 11.23  & 16.28  & 13.56  & 17.11  & 14.47  & 17.43  & 14.86  & 16.73  & 14.02  & \textbf{17.77}  & \textbf{15.08}  \\ 
 \hline
\multirow{4}{*}{image12} & block mask1 & 		36.67  & 30.04  & 38.48  & 32.49  & 39.64  & 33.08  & 39.96  & 33.35  & 38.74  & 32.88  & \textbf{40.23}  & \textbf{33.63 } \\ 
& block mask2 & 		35.87  & 28.23  & 37.37  & 30.74  & 38.01  & 31.41  & 38.40  & 31.80  & 37.65  & 31.04  & \textbf{38.82}  & \textbf{32.22}  \\ 
& block mask3 & 		35.99  & 29.40  & 38.03  & 31.36  & 39.07  & 32.47  & 39.59  & 32.99  & 38.63  & 32.02  & \textbf{39.83}  & \textbf{33.23}  \\
& block mask4 & 		35.73  & 28.12  & 37.95  & 30.42  & 38.93  & 32.34  & 39.35  & 32.76  & 38.52  & 31.91  & \textbf{39.59}  & \textbf{32.99}  \\ 
& block mask5 & 		34.39  & 27.70  & 36.28  & 29.67  & 37.31  & 30.71  & 37.83  & 31.25  & 36.73  & 30.13  & \textbf{38.28}  & \textbf{31.64} \\ 	
		\Xhline{1pt}
	\end{tabular}
		}
\end{center}
	\vspace{-0.2cm}
\end{table*}

\subsection{Convergence Analysis}\label{sec.4.5}

\begin{figure}[!t]
	\centering
	\includegraphics[width=3in]{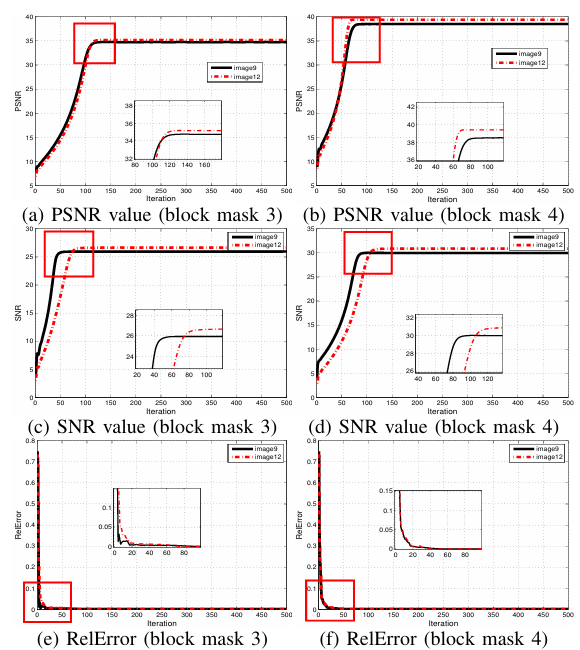}
	\caption{Convergence of the proposed model for different block masks.}\label{fig.10}
	\vspace{-0.2cm}
\end{figure}

In this subsection, to verify the convergence of Algorithm \ref{alg.1}, we select the `image9' and `image12' with the block masks 3 and masks 4 as test images.
Fig. \ref{fig.10} illustrates the functional relationship between the recovered PSNR, SNR, and RE values and the number of iterations.

Observing Fig. \ref{fig.10} reveals that the PSNR and SNR values gradually increase and stabilize as the number of iterations increases, while the RE value decreases rapidly and tends towards zero. Thus, the experimental results validate the convergence of Algorithm \ref{alg.1}. 

Furthermore, to delve deeply into the convergence properties of Algorithm \ref{alg.1}, we conduct a rigorous numerical verification.
We first make the following fundamental assumptions regarding the objective functions in the minimization problem \eqref{eq.b7}.

\begin{Assumption}\label{Ass.1}	
The functions $ R\left( {\bf{W}} \right) $ and $ \phi \left( {\bf{E}} \right) $ are closed and lower semi-continuous within their respective domains, and their gradient functions satisfy the Lipschitz continuity condition. Specifically, there exist positive constants $ {S_1} $ and $ {S_2} $ such that for any $ {x_1} $, $ {x_2} $  in the domain of $ R\left( {\bf{W}} \right) $ and any $ {x_3} $, $ {x_4} $ in the domain of $ \phi \left( {\bf{E}} \right) $, it holds that
\[{\left\| {\nabla R\left( {{x_1}} \right) - \nabla R\left( {{x_2}} \right)} \right\|_F} \le {S_1}{\left\| {{x_1} - {x_2}} \right\|_F},\]	
\[{\left\| {\nabla \phi \left( {{x_3}} \right) - \nabla \phi \left( {{x_4}} \right)} \right\|_F} \le {S_2}{\left\| {{x_3} - {x_4}} \right\|_F}.\]
\end{Assumption}

\begin{Assumption}\label{Ass.2}
The objective function $ R\left( {\bf{W}} \right) + \phi \left( {\bf{E}} \right) $ is bounded below.	
\end{Assumption}

\begin{theorem}\label{th.4}
	Let $ \left\{ {{{\bf{\Gamma }}^k}} \right\}_{k = 1}^{ + \infty }$ be the sequence generated by Algorithm \ref{alg.1}, then there must exist positive constants $ {c_1} $ and $ {c_2} $ such that the following inequality holds,
	\begin{equation}\label{eq.b24}
	\begin{array}{l}
	L\left( {{{\bf{X}}^{k + 1}},\;{{\bf{W}}^k},\;{{\cal E}^k},\;{\bf{F}}_\Omega ^k,\;{{\bf{E}}^k},\;{\bm{\mu }}_1^k,{\bm{\mu }}_2^k,{\bm{\mu }}_3^k} \right)\\
	\; - L\left( {{{\bf{X}}^k},\;{{\bf{W}}^k},\;{{\cal E}^k},\;{\bf{F}}_\Omega ^k,\;{{\bf{E}}^k},\;{\bm{\mu }}_1^k,{\bm{\mu }}_2^k,{\bm{\mu }}_3^k} \right)\\
	\; \le  - {c_1}\left\| {{{\bf{X}}^{k + 1}} - {{\bf{X}}^k}} \right\|_F^2 - {c_2}\left\| {{\bf{X}}_\Omega ^{k + 1} - {\bf{X}}_\Omega ^k} \right\|_F^2.
	\end{array}
	\end{equation}
\end{theorem}

\begin{theorem}\label{th.5}
	Let $ \left\{ {{{\bf{\Gamma }}^k}} \right\}_{k = 1}^{ + \infty }$ be the sequence generated by Algorithm \ref{alg.1}, then there must exist a positive constant $ {c_3} $ such that the following inequality holds,
	\begin{equation}\label{eq.b25}
	\begin{array}{l}
	L\left( {{{\bf{X}}^{k + 1}},\;{{\bf{W}}^{k + 1}},\;{{\cal E}^k},\;{\bf{F}}_\Omega ^k,\;{{\bf{E}}^k},\;{\bm{\mu }}_1^k,{\bm{\mu }}_2^k,{\bm{\mu }}_3^k} \right)\\
	\; - L\left( {{{\bf{X}}^{k + 1}},\;{{\bf{W}}^k},\;{{\cal E}^k},\;{\bf{F}}_\Omega ^k,\;{{\bf{E}}^k},\;{\bm{\mu }}_1^k,{\bm{\mu }}_2^k,{\bm{\mu }}_3^k} \right)\\
	\; \le  - \alpha {c_3}\left\| {{{\bf{W}}^{k + 1}} - {{\bf{W}}^k}} \right\|_F^2.
	\end{array}
	\end{equation}
\end{theorem}

\begin{theorem}\label{th.6}
    Let $ \left\{ {{{\bf{\Gamma }}^k}} \right\}_{k = 1}^{ + \infty }$ be the sequence generated by Algorithm \ref{alg.1}, then there must exist positive constants $ {c_4} $, $ {c_5} $, $ {c_6} $ and $ {c_7} $ such that the following inequality holds,
	\begin{equation}\label{eq.b26}
	\begin{array}{l}
	L\left( {{{\bf{X}}^{k + 1}},\;{{\bf{W}}^{k + 1}},\;{{\cal E}^{k + 1}},\;{\bf{F}}_\Omega ^k,\;{{\bf{E}}^k},\;{\bm{\mu }}_1^k,{\bm{\mu }}_2^k,{\bm{\mu }}_3^k} \right)\;\\
	\; - L\left( {{{\bf{X}}^{k + 1}},\;{{\bf{W}}^{k + 1}},\;{{\cal E}^k},\;{\bf{F}}_\Omega ^k,\;{{\bf{E}}^k},\;{\bm{\mu }}_1^k,{\bm{\mu }}_2^k,{\bm{\mu }}_3^k} \right)\\
	\; \le  - {c_4}\left\| {{{\cal E}^{k + 1}} - {{\cal E}^k}} \right\|_F^2 - {c_5}\left\| {{\cal E}_\Omega ^{k + 1} - {\cal E}_\Omega ^k} \right\|_F^2,
	\end{array}
	\end{equation}
	\begin{equation}\label{eq.b27}
	\begin{array}{l}
	L\left( {{{\bf{X}}^{k + 1}},\;{{\bf{W}}^{k + 1}},\;{{\cal E}^{k + 1}},\;{\bf{F}}_\Omega ^{k + 1},\;{{\bf{E}}^k},\;{\bm{\mu }}_1^k,{\bm{\mu }}_2^k,{\bm{\mu }}_3^k} \right)\\
	\; - L\left( {{{\bf{X}}^{k + 1}},\;{{\bf{W}}^{k + 1}},\;{{\cal E}^{k + 1}},\;{\bf{F}}_\Omega ^k,\;{{\bf{E}}^k},\;{\bm{\mu }}_1^k,{\bm{\mu }}_2^k,{\bm{\mu }}_3^k} \right)\;\\
	\; \le  - {c_6}\left\| {{\bf{F}}_\Omega ^{k + 1} - {\bf{F}}_\Omega ^k} \right\|_F^2,
	\end{array}
	\end{equation}
	\begin{equation}\label{eq.b28}
	\begin{array}{l}
	L\left( {{{\bf{X}}^{k + 1}},\;{{\bf{W}}^{k + 1}},\;{{\cal E}^{k + 1}},\;{\bf{F}}_\Omega ^{k + 1},\;{{\bf{E}}^{k + 1}},\;{\bm{\mu }}_1^k,{\bm{\mu }}_2^k,{\bm{\mu }}_3^k} \right)\;\\
	\; - L\left( {{{\bf{X}}^{k + 1}},\;{{\bf{W}}^{k + 1}},\;{{\cal E}^{k + 1}},\;{\bf{F}}_\Omega ^{k + 1},\;{{\bf{E}}^k},\;{\bm{\mu }}_1^k,{\bm{\mu }}_2^k,{\bm{\mu }}_3^k} \right)\\
	\; \le  - \lambda {c_7}\left\| {{{\bf{E}}^{k + 1}} - {{\bf{E}}^k}} \right\|_F^2.
	\end{array}
	\end{equation}
\end{theorem}

\begin{theorem}\label{th.7}
	Let $ \left\{ {{{\bf{\Gamma }}^k}} \right\}_{k = 1}^{ + \infty }$ be the sequence generated by Algorithm \ref{alg.1}, then there must exist positive constants $ {c_8} $, $ {c_9} $ and $ {c_{10}} $ such that the following inequality holds,
	\begin{equation}\label{eq.b29}
	\begin{array}{l}
	L\left( {{{\bf{X}}^{k + 1}},\;{{\bf{W}}^{k + 1}},\;{{\cal E}^{k + 1}},\;{\bf{F}}_\Omega ^{k + 1},\;{{\bf{E}}^{k + 1}},\;{\bm{\mu }}_1^{k + 1},{\bm{\mu }}_2^{k + 1},{\bm{\mu }}_3^{k + 1}} \right)\;\\
	\; - L\left( {{{\bf{X}}^{k + 1}},\;{{\bf{W}}^{k + 1}},\;{{\cal E}^{k + 1}},\;{\bf{F}}_\Omega ^{k + 1},\;{{\bf{E}}^k},\;{\bm{\mu }}_1^k,{\bm{\mu }}_2^k,{\bm{\mu }}_3^k} \right)\\
	\; \le {c_8}\left\| {{{\bf{W}}^{k + 1}} - {{\bf{W}}^k}} \right\|_F^2 + {c_9}\left\| {{{\bf{E}}^{k + 1}} - {{\bf{E}}^k}} \right\|_F^2\\
	\; + {c_{10}}\left\| {{\bf{F}}_\Omega ^{k + 1} - {\bf{F}}_\Omega ^k} \right\|_F^2.
	\end{array}
	\end{equation}
\end{theorem}

\begin{theorem}\label{th.8}
	Let $ \left\{ {{{\bf{\Gamma }}^k}} \right\}_{k = 1}^{ + \infty }$ be the sequence generated by Algorithm \ref{alg.1}, and the penalty parameter $ \rho $ is large enough, such that when $ \rho  \ge \max \left\{ {\frac{{{\alpha ^2}S_1^2}}{{{c_3}}},\frac{{{\lambda ^2}S_2^2}}{{{c_7}}},\frac{1}{{{c_6}}}} \right\} $, the following inequality holds,
	\begin{equation}\label{eq.b30}
	\begin{array}{l}
	L\left( {{{\bf{X}}^{k + 1}},\;{{\bf{W}}^{k + 1}},\;{{\cal E}^{k + 1}},\;{\bf{F}}_\Omega ^{k + 1},\;{{\bf{E}}^{k + 1}},\;{\bm{\mu }}_1^{k + 1},{\bm{\mu }}_2^{k + 1},{\bm{\mu }}_3^{k + 1}} \right)\;\\
	\;\; \le L\left( {{{\bf{X}}^k},\;{{\bf{W}}^k},\;{{\cal E}^k},\;{\bf{F}}_\Omega ^k,\;{{\bf{E}}^k},\;{\bm{\mu }}_1^k,{\bm{\mu }}_2^k,{\bm{\mu }}_3^k} \right).
	\end{array}
	\end{equation}
\end{theorem}

\begin{theorem}\label{th.9}
	Let $ \left\{ {{{\bf{\Gamma }}^k}} \right\}_{k = 1}^{ + \infty }$ be the sequence generated by Algorithm \ref{alg.1}, then the sequence is bounded.
\end{theorem}

\begin{theorem}\label{th.10}
	Let $ \left\{ {{{\bf{\Gamma }}^k}} \right\}_{k = 1}^{ + \infty }$ be the sequence generated by Algorithm \ref{alg.1}, we have
	\begin{equation}\label{eq.b31}
	\mathop {\lim }\limits_{k \to  + \infty } \left\| {{{\bf{\Gamma }}^{k + 1}} - {{\bf{\Gamma }}^k}} \right\|_F^2 = 0.
	\end{equation}
\end{theorem}

\begin{theorem}\label{th.11}
	Let $ \left\{ {{{\bf{\Gamma }}^k}} \right\}_{k = 1}^{ + \infty }$ be the sequence generated by Algorithm \ref{alg.1}, then any limit point ${{\bf{\Gamma }}^ * }$ of the sequence, which is defined as $ {{\bf{\Gamma }}^ * } = \left( {{{\bf{X}}^ * },\;{{\bf{W}}^ * },\;{{\cal E}^ * },\;{\bf{F}}_\Omega ^ * ,\;{{\bf{E}}^ * },\;{\bm{\mu }}_1^ * ,{\bm{\mu }}_2^ * ,{\bm{\mu }}_3^ * } \right) $, is a stable point of the formula \eqref{eq.b5}.
\end{theorem}

\begin{remark}\label{re.5}
	Based on the above assumptions and Theorems \ref{th.4}-\ref{th.11}, it has been proven that the sequence $ {{\bf{\Gamma }}^k} = L\left( {{{\bf{X}}^k},\;{{\bf{W}}^k},\;{{\cal E}^k},\;{\bf{F}}_\Omega ^k,\;{{\bf{E}}^k},\;{\bm{\mu }}_{_1}^k,\;{\bm{\mu }}_{_2}^k,\;{\bm{\mu }}_{_3}^k} \right) $ generated by Algorithm \ref{alg.1} converges to a stable point $ {{\bf{\Gamma }}^ * } = L\left( {{{\bf{X}}^ * },\;{{\bf{W}}^ * },\;{{\cal E}^ * },\;{\bf{F}}_\Omega ^ * ,\;{{\bf{E}}^ * },\;{\bm{\mu }}_1^ * ,{\bm{\mu }}_2^ * ,{\bm{\mu }}_3^ * } \right) $ of the augmented Lagrangian function $ L\left( {{\bf{X}},\;{\bf{W}},\;{\cal E},\;{{\bf{F}}_\Omega },\;{\bf{E}},\;{{\bm{\mu }}_1},\;{{\bm{\mu }}_2},\;{{\bm{\mu }}_3}} \right) $.
	The detailed proofs of the theorems are provided in Appendices \ref{app.4}-\ref{app.11}. 	
\end{remark}

\section{Conclusion}\label{sec.5}

In this paper, we design a novel low-rank matrix decomposition architecture which assumes that the matrix consists of both low-rank and structured sparse components. On this basis, we propose a generalized low-rank matrix completion model with OGER. In our model, OGER employs the group sparsity metric to quantify the structured sparse elements within matrix data. This enhances the model's capacity to capture structural information, thereby improving its fitting capability and effectively preserving sparse details in the reconstructed matrix. 
Moreover, the proposed model is computationally solved by integrating the MM algorithm into the ADMM framework. To validate the efficacy of OGER, we conduct recovery experiments on images using random masks, text masks, and block masks. The experimental outcomes illustrate that our model surpasses alternative approaches in terms of matrix recovery performance.
It is important to clarify that the current proposed model is designed specifically for low-rank matrix completion. In future research, our aim is to expand its capabilities to include tensor completion as well.

\section*{Acknowledgements}\label{sec.6}

This work was supported in part by the Natural Science Foundation of China under Grant No. 62061016 and 61561019, the Doctoral Scientific Fund Project of Hubei Minzu University under Grant No. MY2015B001, the Internal Scientific Research Project of Hubei Minzu University under Grant No. XN2305, and the Graduate Education Innovation Project of Hubei Minzu University under Grant No. MYK2024006 and STK2023002.

\ifCLASSOPTIONcaptionsoff
\newpage
\fi

\bibliographystyle{IEEEtran}
\bibliography{IEEEabrv,OGERbib}

\begin{IEEEbiography}[{\includegraphics[width=1in,height=1.25in,clip,keepaspectratio]{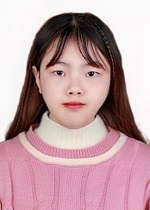}}]{Wenjing Lu}
	was born in 2000. She is currently pursuing her M.S. at Hubei Minzu University. Her main research interests are data analysis and image processing.
\end{IEEEbiography}

\begin{IEEEbiography}[{\includegraphics[width=1in,height=1.25in,clip,keepaspectratio]{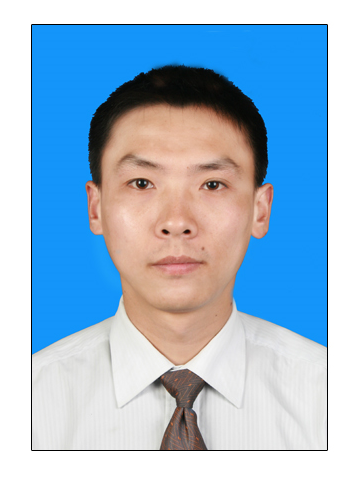}}]{Zhuang Fang}
	was born in 1980. He received the B.S degree from Hubei Minzu University in 2003 and the M.S. degree from Chongqing University in 2011, and Ph.D. degree from Wuhan University in 2019. He is currently an associate professor in the School of Mathematics and Statistics at Hubei Minzu University. His research focuses on image processing, mathematical modeling and low-rank matrix recovery.
\end{IEEEbiography}

\begin{IEEEbiography}[{\includegraphics[width=1in,height=1.25in,clip,keepaspectratio]{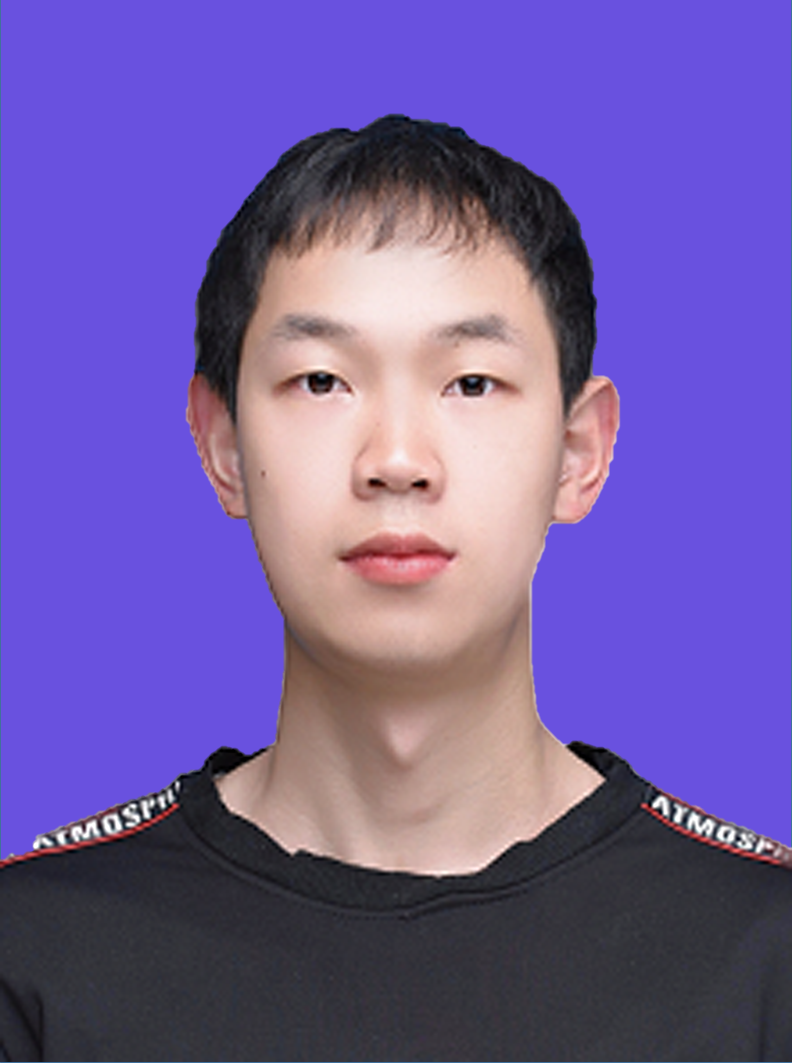}}]{Liang Wu}
	was born in 1998. He recwived the M.S. degree from Hubei Minzu University in 2020. He is currently pursuing a Ph.D. degree with the School of Mathematics and Statistics, Southwest University, Chongqing, China. His current research interests include image processing, low-rank matrix/tensor analysis and high dimensional data modeling.
\end{IEEEbiography}

\begin{IEEEbiography}[{\includegraphics[width=1in,height=1.25in,clip,keepaspectratio]{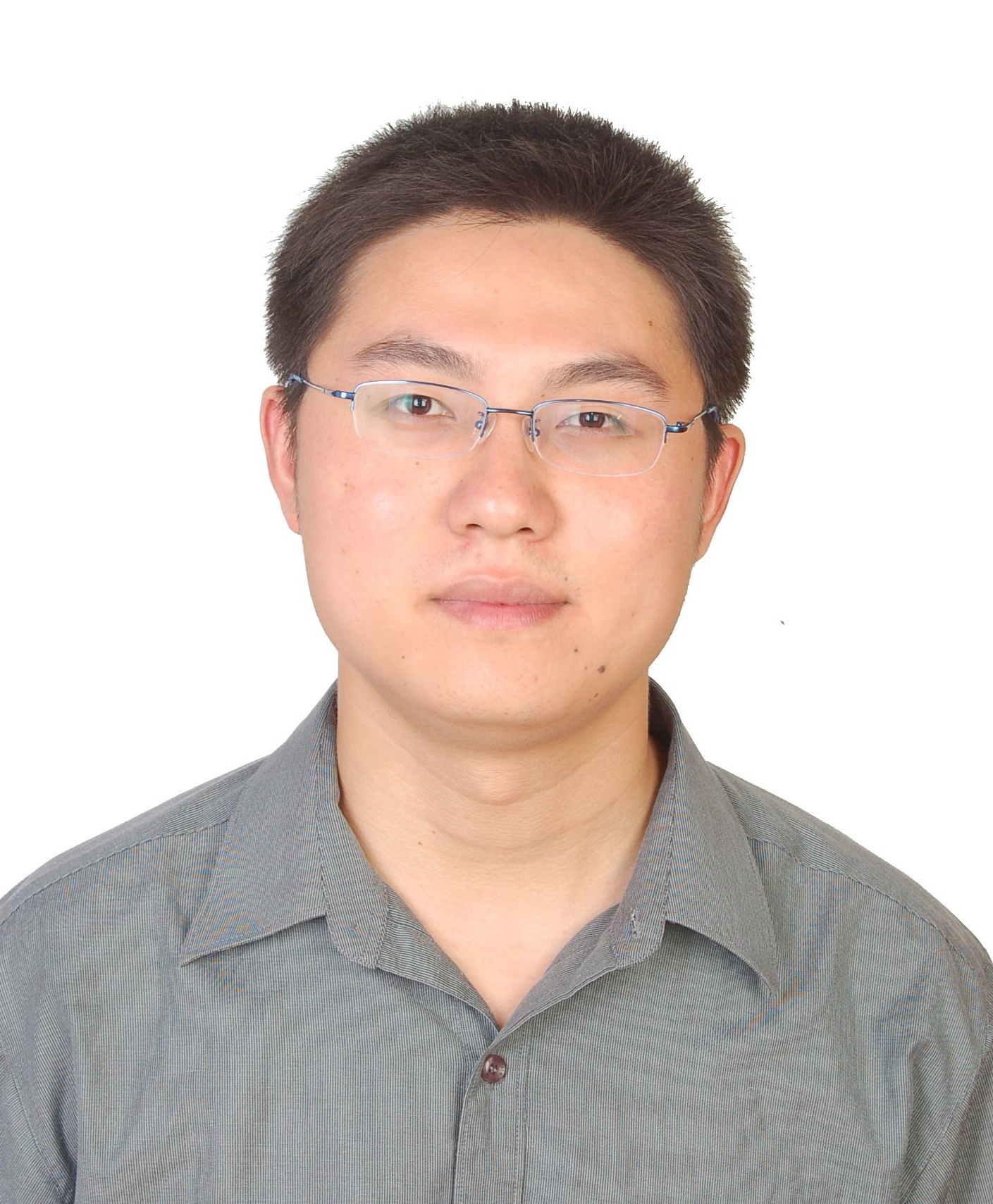}}]{Liming Tang}
	was born in 1978. He received the B.S. degree from Hubei Minzu University in 2000, the M.S. degree from Chongqing University in 2007, and Ph.D. degree in computational mathematics from Chongqing University in 2013. He is currently a professor in the School of Mathematics and Statistics at Hubei Minzu University. His research interests include image processing, computer vision and pattern recognition.
\end{IEEEbiography}

\begin{IEEEbiography}[{\includegraphics[width=1in,height=1.25in,clip,keepaspectratio]{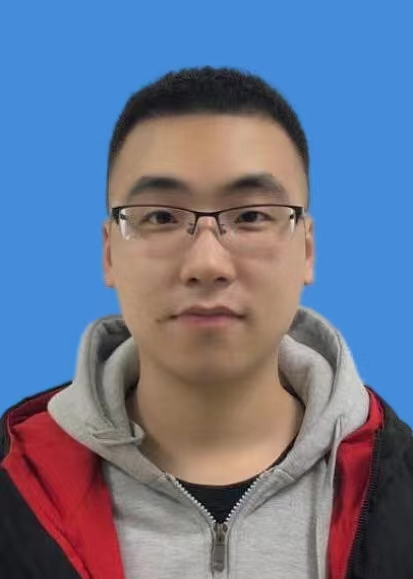}}]{Hanxin Liu}
	was born in 1997. He is currently pursuing his M.S. at Hubei Minzu University. His main research interests are noise estimation and image processing.
\end{IEEEbiography}

\begin{IEEEbiography}[{\includegraphics[width=1in,height=1.25in,clip,keepaspectratio]{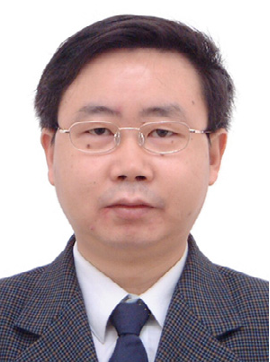}}]{Chuanjiang He}
	received the B.S. and M.S. degree in mathematics from Sichuan University in 1985 and 1988, respectively, and Ph.D. degree from Chongqing University in  2004. He is currently a professor of Mathematics in the School of Mathematics and Statistics at Chongqing University, Chongqing, China. His research interests include partial differential equations and their applications in image processing.
\end{IEEEbiography}

\clearpage

\appendices
\section{PROOF OF THEOREM 1}\label{app.1}

\begin{proof}
The matrix $ {\bf{M}} $ can be decomposition into $ {\bf{M}} = {\bf{U}}{{\bf{\Sigma }}}{{\bf{V}}^T} $, where $ {\bf{U}} $ and $ {\bf{V}} $ are orthogonal matrices, $ {\bf{\Sigma }} $ is a diagonal matrix with diagonal elements $ {\sigma _i} $ and $ r = \min \left\{ {{n_1},{n_2}} \right\} $ is the rank of $ {\bf{M}} $.

Since the matrix $ {{\bf{M}}_k} $ preserves the first $ k $ non-zero singular values in  $ {\bf{\Sigma }} $, we have $ {{\bf{M}}_k} = {\bf{U}}{{\bf{\Sigma }}_k}{{\bf{V}}^T} $, where $ {{\bf{\Sigma }}_k}{\rm{ = diag}}\left( {{\sigma _1},{\sigma _2}, \ldots ,{\sigma _k},0, \ldots ,0} \right) $.

To simplify the calculation, we express the Frobenius norm of the error matrix $ \left\| {{\bf{M}} - {{\bf{M}}_k}} \right\|_F^2 $ as follows,
\begin{equation}\label{eq.c1}
\left\| {{\bf{M}} - {{\bf{M}}_k}} \right\|_F^2 = {\rm{tr}}\left( {{{\left( {{\bf{M}} - {{\bf{M}}_k}} \right)}^T}\left( {{\bf{M}} - {{\bf{M}}_k}} \right)} \right).
\end{equation}

The error matrix $ \left( {{\bf{M}} - {{\bf{M}}_k}} \right) $ can be expanded into the singular value form, that is
\begin{equation}\label{eq.c2}
{\bf{M}} - {{\bf{M}}_k} = \sum\nolimits_{i = k + 1}^r {{\sigma _i}} {u_i}v_i^T,
\end{equation}
where $ {{u_i}} $ and $ {{v_i}} $ are the $ i $-th left and right singular vectors of $ {\bf{M}} $, respectively.

According to the trace operation formula, Eq. \eqref{eq.c1} can be written as
\begin{equation}\label{eq.c3}
\begin{array}{l}
\left\| {{\bf{M}} - {{\bf{M}}_k}} \right\|_F^2 = {\rm{tr}}\left( {{{\left( {\sum\nolimits_{i = k + 1}^r {{\sigma _i}} {u_i}v_i^T} \right)}^T}\left( {\sum\nolimits_{i = k + 1}^r {{\sigma _i}} {u_i}v_i^T} \right)} \right)\\
\;\;\;\;\;\;\;\;\;\;\;\;\;\;\;\;\,\;\;\;\; = {\rm{tr}}\left( {{{\left( {\sum\nolimits_{i = k + 1}^r {{\sigma _i}} {v_i}u_i^T} \right)}^T}\left( {\sum\nolimits_{i = k + 1}^r {{\sigma _i}} {u_i}v_i^T} \right)} \right)\\
\;\;\;\;\;\;\;\;\;\;\;\;\;\;\;\;\,\;\;\;\; = {\rm{tr}}\left( {\sum\nolimits_{i = k + 1}^r {{\sigma _i}} {v_i}v_i^T} \right)\\
\;\;\;\;\;\;\;\;\;\;\;\;\;\;\;\;\,\;\;\;\; = \sum\nolimits_{i = k + 1}^r {\sigma _i^2} {\rm{tr}}\left( {{v_i}v_i^T} \right).
\end{array}
\end{equation}

Therefore, we can obtain
\begin{equation}\label{eq.c4}
\left\| {{\bf{M}} - {{\bf{M}}_k}} \right\|_F^2 = \sum\nolimits_{i = k + 1}^r {\sigma _i^2}. 
\end{equation}
Then, $ \left\| {{\bf{M}} - {{\bf{M}}_k}} \right\|_F^2 $ can be expressed by the singular value $ {\sigma _i} $ of $ {\bf{M}} $.
	
Thus, Theorem \ref{th.1} is obtained.
\end{proof}

\section{PROOF OF THEOREM 2}\label{app.2}

\begin{proof}
	To prove Theorem \ref{th.2}, we first define some terms.
	Given two matrices $ {{\cal E}_1}  \in\mathbb{R}^{n_1\times n_2} $ and $ {{\cal E}_2}  \in\mathbb{R}^{n_1\times n_2} $, where $ {{\cal E}_1}_{\left( {i,j} \right),K} $ and $ {{\cal E}_2}_{\left( {i,j} \right),K} $ represent points groups of size $ K*K $ centered on $ \left( {i,j} \right) $.
	
	According to the definition of convex function, we need to prove that for any $ {{\cal E}_1} $,  $ {{\cal E}_2} $, and $ \theta  \in \left[ {0,1} \right] $, it holds that
	\begin{equation}\label{eq.c5}
	\phi \left( {\theta {{\cal E}_1} + \left( {1 - \theta } \right){{\cal E}_2}} \right) \le \theta \phi \left( {{{\cal E}_1}} \right) + \left( {1 - \theta } \right)\phi \left( {{{\cal E}_2}} \right). 
	\end{equation}
	Then the function $ \phi \left( {\cal E} \right) $ is a convex function.
	
	By the definition of $ \phi \left( {\cal E} \right) $, we expand the term $ \phi \left( {\theta {{\cal E}_1} + \left( {1 - \theta } \right){{\cal E}_2}} \right) $ on the left side of inequality \eqref{eq.c5}, for each $ \left( {i,j} \right) $, we have
	\begin{equation}\label{eq.c6}
	\resizebox{0.89\linewidth}{!}{$
	\sum\nolimits_{i,j = 1} {{{\left[ {{{\sum\nolimits_{{k_1},{k_2} =  - {n_1}}^{{n_2}} {\left| {\theta {\varepsilon _1}\left( {i + {k_1},j + {k_2}} \right) + \left( {1 - \theta } \right){\varepsilon _2}\left( {i + {k_1},j + {k_2}} \right)} \right|} }^2}} \right]}^{\frac{1}{2}}}},
		$}
	\end{equation}
	where $  {\varepsilon _1} $ and $  {\varepsilon _2} $ are elements of $ {{\cal E}_1} $ and  $ {{\cal E}_2} $, respectively.

	By the property of absolute value inequality, for any $ a,b,\theta  \in\mathbb{R} $, there is
	\begin{equation}\label{eq.c7}
	\left| {\theta a + \left( {1 - \theta } \right)b} \right| \le \theta \left| a \right| + \left( {1 - \theta } \right)\left| b \right|.
	\end{equation}
	Applying this property to inequality \eqref{eq.c5}, we have
	\begin{equation}\label{eq.c8}
		\resizebox{0.89\linewidth}{!}{$
	\begin{array}{l}
		\left| {\theta {\varepsilon _1}\left( {i + {k_1},j + {k_2}} \right) + \left( {1 - \theta } \right){\varepsilon _2}\left( {i + {k_1},j + {k_2}} \right)} \right|\\
		\le \theta \left| {{\varepsilon _1}\left( {i + {k_1},j + {k_2}} \right)} \right| + \left( {1 - \theta } \right)\left| {{\varepsilon _2}\left( {i + {k_1},j + {k_2}} \right)} \right|.
	\end{array}
		$}
	\end{equation}
	Since the square function is convex, we square both sides of the inequality, that is
	\begin{equation}\label{eq.c9}
	\resizebox{0.89\linewidth}{!}{$
	\begin{array}{l}
	{\left( {\theta {\varepsilon _1}\left( {i + {k_1},j + {k_2}} \right) + \left( {1 - \theta } \right){\varepsilon _2}\left( {i + {k_1},j + {k_2}} \right)} \right)^2}\\
	\le \theta {\left| {{\varepsilon _1}\left( {i + {k_1},j + {k_2}} \right)} \right|^2} + \left( {1 - \theta } \right){\left| {{\varepsilon _2}\left( {i + {k_1},j + {k_2}} \right)} \right|^2}.
	\end{array}
	$}
	\end{equation}
	By summing up the above inequality for all $ i,j,{k_1},{k_2} $, we get
	\begin{equation}\label{eq.c10}
	\phi \left( {\theta {{\cal E}_1} + \left( {1 - \theta } \right){{\cal E}_2}} \right) \le \theta \phi \left( {{{\cal E}_1}} \right) + \left( {1 - \theta } \right)\phi \left( {{{\cal E}_2}} \right). 
	\end{equation}

	By the definition of convex functions, we have proved that $ \phi \left( {\cal E} \right) $ is a convex function. 

	Thus, Theorem \ref{th.2} is obtained.
\end{proof}
	
\section{PROOF OF THEOREM 3}\label{app.3}
\begin{proof}
The matrices $ \bf{W} $ and $ \bf{D} $ are subjected to singular value decomposition to obtain $ {\bf{W}} = {{\bf{U}}_{\bf{W}}}{{\bf{\Sigma }}_{\bf{W}}}{\bf{V}}_{\bf{W}}^{T} $ and $ {\bf{D}} = {{\bf{Q}}_{\bf{D}}}{{\bf{S}}_{\bf{D}}}{\bf{R}}_{\bf{D}}^{T} $,  where $ {{\bf{\Sigma }}_{\bf{W}}} $ and $ {{\bf{S}}_{\bf{D}}} $ are the diagonal matrices of $ \bf{W} $ and $ \bf{D} $, respectively.
Consequently, problem \eqref{eq.b12} can be rewritten as
\begin{equation}\label{eq.d1}
\resizebox{0.88\linewidth}{!}{$
\lambda \sum\nolimits_{i = 1}^{\min \left\{ {{n_1},{n_2}} \right\}} {\psi \left( {{\sigma _i}} \right)}  + \frac{\rho }{2}\left\| {{{\bf{U}}_{\bf{W}}}{{\bf{\Sigma }}_{\bf{W}}}{\bf{V}}_{\bf{W}}^{T} - {{\bf{Q}}_{\bf{D}}}{{\bf{S}}_{\bf{D}}}{\bf{R}}_{\bf{D}}^{T}} \right\|_F^2.
	$}
\end{equation}
According to Lemma \ref{le.1}, it follows that
\begin{equation}\label{eq.d2}
\begin{array}{l}
\left\| {{{\bf{U}}_{\bf{W}}}{{\bf{\Sigma }}_{\bf{W}}}{\bf{V}}_{\bf{W}}^{T} - {{\bf{Q}}_{\bf{D}}}{{\bf{S}}_{\bf{D}}}{\bf{R}}_{\bf{D}}^{T}} \right\|_F^2\\
= tr\left( {{\bf{\Sigma }}_{\bf{W}}^{T}{{\bf{\Sigma }}_{\bf{W}}}} \right) + tr\left( {{\bf{S}}_{\bf{D}}^{T}{{\bf{S}}_{\bf{D}}}} \right) - 2tr\left( {{{\bf{W}}^{T}}{\bf{D}}} \right)\\
\ge tr\left( {{\bf{\Sigma }}_{\bf{W}}^{T}{{\bf{\Sigma }}_{\bf{W}}}} \right) + tr\left( {{\bf{S}}_{\bf{D}}^{T}{{\bf{S}}_{\bf{D}}}} \right) - 2tr\left( {{\bf{\Sigma }}_{\bf{W}}^{T}{{\bf{S}}_{\bf{D}}}} \right)\\
= \left\| {{\bf{\Sigma }}_{\bf{W}}^{T}{{\bf{S}}_{\bf{D}}}} \right\|_F^2 \\
= \sum\nolimits_{i = 1}^{\min \left\{ {{n_1},{n_2}} \right\}} {{{\left( {{\sigma _i} - {s_i}} \right)}^2}}, 
\end{array}
\end{equation}
where the equation in the above formula holds if and only if $ {{\bf{U}}_{\bf{W}}} = {{\bf{Q}}_{\bf{D}}} $, $ {{\bf{V}}_{\bf{W}}} = {{\bf{R}}_{\bf{D}}} $.

Therefore, the optimal numerical solution of problem \eqref{eq.d1} can be written as
\begin{equation}\label{eq.d3}
\mathop {\min }\limits_{{\sigma _i}} \lambda \sum\nolimits_{i = 1}^{\min \left\{ {{n_1},{n_2}} \right\}} {\psi \left( {{\sigma _i}} \right)}  + \frac{\rho }{2}{\sum\nolimits_{i = 1}^{\min \left\{ {{n_1},{n_2}} \right\}} {\left( {{\sigma _i} - {s_i}} \right)} ^2}.
\end{equation}
Simplification of the above equation yields that
\begin{equation}\label{eq.d4}
\mathop {\min }\limits_{{\sigma _i}} \lambda \psi \left( {{\sigma _i}} \right) + \frac{\rho }{2}{\left( {{\sigma _i} - {s_i}} \right)^2}.
\end{equation}

Thus, Theorem \ref{th.3} is obtained.
\end{proof}

\section{SOLVE THE $ \bf{E} $-SUBPROBLEM}\label{app.a1}

The $ \bf{E} $-subproblem \eqref{eq.b19} is expressed as
\begin{equation}\label{eq.d5}
{{\bf{E}}^{k + 1}} = \mathop {\arg \min }\limits_{\bf{E}} \lambda \phi \left( {\bf{E}} \right) + \frac{\rho }{2}\left\| {{\bf{E}} - {{\bf{\cal E }}^{k + 1}} + \frac{{{\bm{\mu }}_2^k}}{\rho }} \right\|_F^2.
\end{equation}
For convenience, we reformulate the above equation as
\begin{equation}\label{eq.d6}
{{\bf{E}}^{k + 1}} = \mathop {\arg \min }\limits_{\bf{E}} \left\{ {J\left( {\bf{E}} \right)\; = \lambda \phi \left( {\bf{E}} \right) + \frac{\rho }{2}\left\| {{\bf{E}} -  {{\bf{D}}^k}} \right\|_F^2} \right\},
\end{equation}
where $ {{\bf{D}}^k} = {{\bf{\cal E }}^{k + 1}} + \frac{{{\bm{\mu }}_2^k}}{\rho } $ . It can be seen from the mean inequality that, for all $ \forall {\bf{V}} \in {R^{\dim \left( {\bf{E}} \right)}} $, we have the following,
\begin{equation}\label{eq.d7}
\begin{array}{*{20}{l}}
{\phi \left( {\bf{E}} \right) = \sum\limits_{i,j = 1} {{{\left\| {{{\bf{E}}_{\left( {i,j} \right),K}}} \right\|}_2}} }\\
{\;\;\;\;\;\;\;\;\; \le \sum\limits_{i,j = 1} {\left( {\frac{{\left\| {{{\bf{E}}_{\left( {i,j} \right),K}}} \right\|_2^2}}{{2{{\left\| {{{\bf{V}}_{\left( {i,j} \right),K}}} \right\|}_2}}} + \frac{1}{2}{{\left\| {{{\bf{V}}_{\left( {i,j} \right),K}}} \right\|}_2}} \right)} }\\
{\;\;\;\;\;\;\;\;\; = S\left( {{\bf{E}},{\bf{V}}} \right)}.
\end{array}
\end{equation}

Thus, we claim that $ T\left( {{\bf{E}},{\bf{V}}} \right) = \lambda S\left( {{\bf{E}},{\bf{V}}} \right) + \frac{\rho }{2}\left\| {{\bf{E}} -  {{\bf{D}}^k}} \right\|_F^2 $ is a substitute function of the function $ \phi \left( {\bf{E}} \right) $ at point $ \bf{V} $. In addition,  there are two facts here:
(i) $ T\left( {{\bf{E}},{\bf{V}}} \right) \ge J\left( {\bf{E}} \right) $ and (ii) $ T\left( {{\bf{V}},{\bf{V}}} \right) = J\left( {\bf{V}} \right) $. Furthermore, by Theorem \ref{de.8}, we may as well take $ {\bf{V}} = {{\bf{E}}^k} $ and convert the $ \bf{E} $-subproblem into
\begin{equation}\label{eq.d8}
\begin{array}{l}
{{\bf{E}}^{k + 1}} = \mathop {\arg \min }\limits_{\bf{E}} T\left( {{\bf{E}},{{\bf{E}}^k}} \right)\\
\;= \mathop {\arg \min }\limits_{\bf{E}} \frac{\rho }{2}\left\| {{\bf{E}} - {{\bf{D}}^k}} \right\|_F^2 + \frac{\lambda }{2}\left\| {{\bf{\Lambda }}\left( {{{\bf{E}}^k}} \right){\bf{E}}} \right\|_2^2 +{\bf{C}},
\end{array}
\end{equation}
where $ {\bf{C}} $  is a constant independent of the variable $ \bf{E} $ and $ {\bf{\Lambda }}\left( {{{\bf{E}}^k}} \right) $ is a diagonal matrix whose elements on the diagonal can be expressed as
\[
\resizebox{\linewidth}{!}{$
	{\left[ {{\bf{\Lambda }}\left( {{{\bf{E}}^k}} \right)} \right]_{l,l}} = \sqrt {{{\sum\limits_{i,j =  - {m_1}}^{{m_2}} {\left[ {\sum\limits_{{k_1},{k_2} =  - {m_1}}^{{m_2}} {{{\left| {{{\bf{E}}^k}_{r - i + {k_1},i - j + {k_2}}} \right|}^2}} } \right]} }^{ - 1/2}}}.
	$}
\]

Finally, we find the optimized solution of the $ \bf{E} $-subproblem as
\begin{equation}\label{eq.d9}
{{\bf{E}}^{k + 1}} = {\left( {{\bf{I}} + \frac{{\lambda {\bf{\Lambda }}{{\left( {{{\bf{E}}^k}} \right)}^T}{\bf{\Lambda }}\left( {{{\bf{E}}^k}} \right)}}{\rho }} \right)^{ - 1}}{{\bf{D}}^k}.
\end{equation}

\section{PROOF OF THEOREM 4}\label{app.4}
\begin{proof}
	Firstly, we prove that the function $ L $ is strictly convex with respect to the $ {\bf{X}} $- subproblem.
	
	Expanding the first term of Eq. \eqref{eq.b7}, we have
	\begin{flalign}\label{eq.e1}
	\begin{split}
	\left\| {{\bf{W}} - {\bf{X}} + \frac{{{{\bm{\mu }}_1}}}{\rho }} \right\|_F^2 & = \left\| {\bf{X}} \right\|_F^2 + \left\langle {{\bf{X}}, - {\bf{W}} - \frac{{{{\bm{\mu }}_1}}}{\rho }} \right\rangle \\
	&+ \left\| { - {\bf{W}} - \frac{{{{\bm{\mu }}_1}}}{\rho }} \right\|_F^2.
	\end{split}&
	\end{flalign}
	Considering that this term is a linear term with respect to $ {\bf{X}} $, i.e., it is convex.
	
	Next, expanding the second term of Eq. \eqref{eq.b7}, we get
	\begin{equation}\label{eq.e2}
	\begin{array}{l}
	\left\| {{{\bf{F}}_\Omega } - \left( {{{\bf{Y}}_\Omega } - {{\bf{X}}_\Omega } - {{\cal E}_\Omega }} \right) + \frac{{{{\bm{\mu }}_3}}}{\rho }} \right\|_F^2\\
	\; = \left\| {\bf{X}} \right\|_F^2 + \left\langle {{\bf{X}},{{\bf{F}}_\Omega } - \left( {{{\bf{Y}}_\Omega } - {{\cal E}_\Omega }} \right) + \frac{{{{\bm{\mu }}_3}}}{\rho }} \right\rangle \\
	\; + \left\| {{{\bf{F}}_\Omega } - \left( {{{\bf{Y}}_\Omega } - {{\cal E}_\Omega }} \right) + \frac{{{{\bm{\mu }}_3}}}{\rho }} \right\|_F^2
	\end{array}
	\end{equation}
	It can be seen that this term is also a linear term with respect to $ {\bf{X}} $, meaning it is also convex.
	
	Therefore, it is demonstrated that the function $ L $ is strictly convex with respect to the $ {\bf{X}} $-subproblem.
	
	The Taylor expansion and the properties of convex functions \cite{yang2013linearized} indicate that
	\begin{equation}\label{eq.e3}
	f\left( {{{\bf{X}}^{k + 1}}} \right) \ge f\left( {{{\bf{X}}^k}} \right) + \left\langle {\nabla f\left( {{{\bf{X}}^k}} \right),{{\bf{X}}^{k + 1}} - {{\bf{X}}^k}} \right\rangle.
	\end{equation}
	Consequently, we can deduce that there must exist positive constants $ {c_1} $ and $ {c_2} $ such that the following inequality holds,
	\begin{equation}\label{eq.e4}
	\begin{array}{l}
	L\left( {{{\bf{X}}^{k + 1}},\;{{\bf{W}}^k},\;{{\cal E}^k},\;{\bf{F}}_\Omega ^k,\;{{\bf{E}}^k},\;{\bm{\mu }}_1^k,{\bm{\mu }}_2^k,{\bm{\mu }}_3^k} \right)\;\\
	\; - L\left( {{{\bf{X}}^k},\;{{\bf{W}}^k},\;{{\cal E}^k},\;{\bf{F}}_\Omega ^k,\;{{\bf{E}}^k},\;{\bm{\mu }}_1^k,{\bm{\mu }}_2^k,{\bm{\mu }}_3^k} \right)\\
	\; \le  - {c_1}\left\| {{{\bf{X}}^{k + 1}} - {{\bf{X}}^k}} \right\|_F^2 - {c_2}\left\| {{\bf{X}}_\Omega ^{k + 1} - {\bf{X}}_\Omega ^k} \right\|_F^2\\
	\; + \left\langle {{\nabla _{\bf{X}}}{\rm I}\left( {{{\bf{X}}^{k + 1}}} \right),{{\bf{X}}^{k + 1}} - {{\bf{X}}^k}} \right\rangle, 
	\end{array}
	\end{equation}
	where $ {\rm I}\left( {{{\bf{X}}^{k + 1}}} \right) = L\left( {{{\bf{X}}^{k + 1}},\;{{\bf{W}}^k},\;{{\cal E}^k},\;{\bf{F}}_\Omega ^k,\;{{\bf{E}}^k},\;{\bm{\mu }}_1^k,{\bm{\mu }}_2^k,{\bm{\mu }}_3^k} \right)$ represents the gradient of the function $ L $ with respect to $ {\bf{X}} $ at point $ {{{\bf{X}}^{k + 1}}} $. 
	Since $ {{{\bf{X}}^{k + 1}}} $ is the point at which $ L $ reaches a local minimum, we have
	\begin{equation}\label{eq.e5}
	{\nabla _{\bf{X}}}L\left( {{{\bf{X}}^{k + 1}},\;{{\bf{W}}^k},\;{{\cal E}^k},\;{\bf{F}}_\Omega ^k,\;{{\bf{E}}^k},\;{\bm{\mu }}_1^k,{\bm{\mu }}_2^k,{\bm{\mu }}_3^k} \right) = 0.
	\end{equation}
	Combining \eqref{eq.e4} and \eqref{eq.e5}, we can obtain
	\begin{equation}\label{eq.e6}
	\begin{array}{l}
	L\left( {{{\bf{X}}^{k + 1}},\;{{\bf{W}}^k},\;{{\cal E}^k},\;{\bf{F}}_\Omega ^k,\;{{\bf{E}}^k},\;{\bm{\mu }}_1^k,{\bm{\mu }}_2^k,{\bm{\mu }}_3^k} \right)\\
	\; - L\left( {{{\bf{X}}^k},\;{{\bf{W}}^k},\;{{\cal E}^k},\;{\bf{F}}_\Omega ^k,\;{{\bf{E}}^k},\;{\bm{\mu }}_1^k,{\bm{\mu }}_2^k,{\bm{\mu }}_3^k} \right)\\
	\; \le  - {c_1}\left\| {{{\bf{X}}^{k + 1}} - {{\bf{X}}^k}} \right\|_F^2 - {c_2}\left\| {{\bf{X}}_\Omega ^{k + 1} - {\bf{X}}_\Omega ^k} \right\|_F^2.
	\end{array}
	\end{equation}	
	
	Thus, Theorem \ref{th.4} is obtained.
\end{proof}	

\section{PROOF OF THEOREM 5}\label{app.5}
\begin{proof}
	Considering that $ R\left( {\bf{W}} \right) $ is a generalized approximation function of a rank function, it is necessary to discuss it separately as a convex and non-convex approximation function.
	
	(i) If $ R\left( {\bf{W}} \right) $ is a convex approximation function, we can determine from \eqref{eq.b10} that function $ L $ is strictly convex with respect to the $ {\bf{W}} $-subproblem.
	 
	Similar to Theorem \ref{th.4}, we can deduce that there must exist a positive constant $ {c_3} $ such that the following inequality holds,
	\begin{equation}\label{eq.f1}
	\begin{array}{l}
	L\left( {{{\bf{X}}^{k + 1}},\;{{\bf{W}}^{k + 1}},\;{{\cal E}^k},\;{\bf{F}}_\Omega ^k,\;{{\bf{E}}^k},\;{\bm{\mu }}_1^k,{\bm{\mu }}_2^k,{\bm{\mu }}_3^k} \right)\\
	\; - L\left( {{{\bf{X}}^{k + 1}},\;{{\bf{W}}^k},\;{{\cal E}^k},\;{\bf{F}}_\Omega ^k,\;{{\bf{E}}^k},\;{\bm{\mu }}_1^k,{\bm{\mu }}_2^k,{\bm{\mu }}_3^k} \right)\\
	\; \le  - \alpha {c_3}\left\| {{{\bf{W}}^{k + 1}} - {{\bf{W}}^k}} \right\|_F^2.
	\end{array}
	\end{equation}	
	
	(ii)  If $ R\left( {\bf{W}} \right) $ is a non-convex approximation function, then since $ {{\bf{W}}^{k+1}} $ is a minimum in the minimization problem \eqref{eq.b10}, we have
	\begin{equation}\label{eq.f2}
	\begin{array}{l}
	\alpha \left\langle {\nabla R\left( {{{\bf{W}}^k}} \right),{{\bf{W}}^{k + 1}} - {{\bf{W}}^k}} \right\rangle \\
	\; + \frac{\rho }{2}\left\| {{{\bf{W}}^{k + 1}} - {{\bf{X}}^{k + 1}} + \frac{{{\bm{\mu }}_1^k}}{\rho }} \right\|_F^2\\
	\; \le \frac{\rho }{2}\left\| {{{\bf{W}}^k} - {{\bf{X}}^{k + 1}} + \frac{{{\bm{\mu }}_1^k}}{\rho }} \right\|_F^2.
	\end{array}
	\end{equation}
	
	It follows from Assumption \ref{Ass.1} and the properties of non-convex functions \cite{zhang2024singular,zhang2024accelerated} that there must exist a positive constant $ {c_3} $ such that
	\begin{flalign}\label{eq.f3}
	\begin{split}
	R\left( {{{\bf{W}}^{k + 1}}} \right)& \le R\left( {{{\bf{W}}^k}} \right) + \left\langle {\nabla R\left( {{{\bf{W}}^k}} \right),{{\bf{W}}^{k + 1}} - {{\bf{W}}^k}} \right\rangle \\
	&- {c_3}\left\| {{{\bf{W}}^{k + 1}} - {{\bf{W}}^k}} \right\|_F^2.
	\end{split}&
	\end{flalign}

	Combining \eqref{eq.f2} and \eqref{eq.f3}, we obtain
	\begin{equation}\label{eq.f4}
	\begin{array}{l}
	\alpha R\left( {{{\bf{W}}^{k + 1}}} \right) + \frac{\rho }{2}\left\| {{{\bf{W}}^{k + 1}} - {{\bf{X}}^{k + 1}} + \frac{{{\bm{\mu }}_1^k}}{\rho }} \right\|_F^2\\
	\; \le \alpha R\left( {{{\bf{W}}^k}} \right) + \frac{\rho }{2}\left\| {{{\bf{W}}^k} - {{\bf{X}}^{k + 1}} + \frac{{{\bm{\mu }}_1^k}}{\rho }} \right\|_F^2.
	\end{array}
	\end{equation}
	This implies that
	\begin{equation}\label{eq.f5}
	\begin{array}{l}
	L\left( {{{\bf{X}}^{k + 1}},\;{{\bf{W}}^{k + 1}},\;{{\cal E}^k},\;{\bf{F}}_\Omega ^k,\;{{\bf{E}}^k},\;{\bm{\mu }}_1^k,{\bm{\mu }}_2^k,{\bm{\mu }}_3^k} \right)\\
	\; - L\left( {{{\bf{X}}^{k + 1}},\;{{\bf{W}}^k},\;{{\cal E}^k},\;{\bf{F}}_\Omega ^k,\;{{\bf{E}}^k},\;{\bm{\mu }}_1^k,{\bm{\mu }}_2^k,{\bm{\mu }}_3^k} \right)\\
	\; \le  - \alpha {c_3}\left\| {{{\bf{W}}^{k + 1}} - {{\bf{W}}^k}} \right\|_F^2.
	\end{array}
	\end{equation}	
	
	By (i) and (ii), Theorem \ref{th.5} is obtained.
\end{proof}	

\section{PROOF OF THEOREM 6}\label{app.6}
\begin{proof}
	(i) We can determine from \eqref{eq.b14} that function $ L $ is strictly convex with respect to the $ {\cal E} $-subproblem.
	
	Similar to Theorem \ref{th.4}, we can deduce that there must exist positive constants $ {c_4} $ and $ {c_5} $ such that the following inequality holds,
	\begin{equation}\label{eq.g1}
	\begin{array}{l}
	L\left( {{{\bf{X}}^{k + 1}},\;{{\bf{W}}^{k + 1}},\;{{\cal E}^{k + 1}},\;{\bf{F}}_\Omega ^k,\;{{\bf{E}}^k},\;{\bm{\mu }}_1^k,{\bm{\mu }}_2^k,{\bm{\mu }}_3^k} \right)\;\\
	\; - L\left( {{{\bf{X}}^{k + 1}},\;{{\bf{W}}^{k + 1}},\;{{\cal E}^k},\;{\bf{F}}_\Omega ^k,\;{{\bf{E}}^k},\;{\bm{\mu }}_1^k,{\bm{\mu }}_2^k,{\bm{\mu }}_3^k} \right)\\
	\; \le  - {c_4}\left\| {{{\cal E}^{k + 1}} - {{\cal E}^k}} \right\|_F^2 - {c_5}\left\| {{\cal E}_\Omega ^{k + 1} - {\cal E}_\Omega ^k} \right\|_F^2.
	\end{array}
	\end{equation}

	(ii) Similarly, we can determine from \eqref{eq.b17} that function $ L $ is strictly convex with respect to the $ {{\bf{F}}_\Omega } $-subproblem. 
	
	Therefore, we can deduce that there exists a positive constant $ {c_6} $ such that the following inequality holds,
	\begin{equation}\label{eq.g2}
	\begin{array}{l}
	L\left( {{{\bf{X}}^{k + 1}},\;{{\bf{W}}^{k + 1}},\;{{\cal E}^{k + 1}},\;{\bf{F}}_\Omega ^{k + 1},\;{{\bf{E}}^k},\;{\bf{\mu }}_1^k,{\bf{\mu }}_2^k,{\bf{\mu }}_3^k} \right)\\
	\; - L\left( {{{\bf{X}}^{k + 1}},\;{{\bf{W}}^{k + 1}},\;{{\cal E}^{k + 1}},\;{\bf{F}}_\Omega ^k,\;{{\bf{E}}^k},\;{\bf{\mu }}_1^k,{\bf{\mu }}_2^k,{\bf{\mu }}_3^k} \right)\;\\
	\, \le  - {c_6}\left\| {{\bf{F}}_\Omega ^{k + 1} - {\bf{F}}_\Omega ^k} \right\|_F^2
	\end{array}
	\end{equation}

	(iii) Considering that we have proved that $ \phi \left( \cdot  \right)\ $ is a convex function in Theorem \ref{th.2}. 
	
	Similarly, we can determine from \eqref{eq.b19} that function $ L $ is strictly convex with respect to the $ {\bf{E}}$-subproblem. 
	 
	Therefore, we can deduce that there exists a positive constant $ {c_7} $ such that the following inequality holds,
	\begin{equation}\label{eq.g3}
	\begin{array}{l}
	L\left( {{{\bf{X}}^{k + 1}},\;{{\bf{W}}^{k + 1}},\;{{\cal E}^{k + 1}},\;{\bf{F}}_\Omega ^{k + 1},\;{{\bf{E}}^{k + 1}},\;{\bf{\mu }}_1^k,{\bf{\mu }}_2^k,{\bf{\mu }}_3^k} \right)\;\\
	\; - L\left( {{{\bf{X}}^{k + 1}},\;{{\bf{W}}^{k + 1}},\;{{\cal E}^{k + 1}},\;{\bf{F}}_\Omega ^{k + 1},\;{{\bf{E}}^k},\;{\bf{\mu }}_1^k,{\bf{\mu }}_2^k,{\bf{\mu }}_3^k} \right)\\
	\; \le  - \lambda {c_7}\left\| {{{\bf{E}}^{k + 1}} - {{\bf{E}}^k}} \right\|_F^2.
	\end{array}
	\end{equation}

	Based on (i), (ii) and (iii), Theorem \ref{th.6} is obtained.
\end{proof}	

\section{PROOF OF THEOREM 7}\label{app.7}
\begin{proof}
	Considering that $ {{\bf{W}}^{k + 1}} $, $ 	{\bf{F}}_\Omega ^{k + 1} $, and $ {{\bf{E}}^{k + 1}} $ are minima in the minimization problems \eqref{eq.b10}, \eqref{eq.b17}, and \eqref{eq.b19}, respectively, we have
	\begin{equation}\label{eq.h1}
	\alpha \nabla R\left( {{{\bf{W}}^{k + 1}}} \right) + \rho \left( {{{\bf{W}}^{k + 1}} - {{\bf{X}}^{k + 1}} + \frac{{{\bm{\mu }}_1^k}}{\rho }} \right) = 0,
	\end{equation}
	\begin{equation}\label{eq.h2}
	\lambda \nabla \phi \left( {{{\bf{E}}^{k + 1}}} \right) + \rho \left( {{{\bf{E}}^{k + 1}} - {{\cal E}^{k + 1}} + \frac{{{\bm{\mu }}_2^k}}{\rho }} \right) = 0,
	\end{equation}
	\begin{equation}\label{eq.h3}
	{\bf{F}}_\Omega ^{k + 1} + \rho \left( {{\bf{F}}_\Omega ^{k + 1} - \left( {{{\bf{Y}}_\Omega } - {\bf{X}}_\Omega ^{k + 1} - {\cal E}_\Omega ^{k + 1}} \right) + \frac{{{\bm{\mu }}_3^k}}{\rho }} \right) = 0.
	\end{equation}
	
	Combining \eqref{eq.b21} through \eqref{eq.b23}, we obtain
	\begin{equation}\label{eq.h4}
	{\bm{\mu }}_1^{k + 1} + \alpha \nabla R\left( {{{\bf{W}}^{k + 1}}} \right) = 0,
	\end{equation}
	\begin{equation}\label{eq.h5}
	{\bm{\mu }}_2^{k + 1} + \lambda \nabla \phi \left( {{{\bf{E}}^{k + 1}}} \right) = 0,
	\end{equation}
	\begin{equation}\label{eq.h6}
	{\bm{\mu }}_3^{k + 1} + {\bf{F}}_\Omega ^{k + 1} = 0.
	\end{equation}
	
	Further, by Assumption \ref{Ass.1}, we can deduce that there must exist positive constants $ {c_8} $ , $ {c_9} $ and $ {c_{10}} $ such that the following inequality holds,

	\begin{align}   
	\label{eq.h8}  
	\left\| {{\bm{\mu }}_1^{k + 1} - {\bm{\mu }}_1^k} \right\|_F^2 &= \left\| { - \alpha \nabla R\left( {{{\bf{W}}^{k + 1}}} \right) + \alpha \nabla R\left( {{{\bf{W}}^k}} \right)} \right\|_F^2 \nonumber  \\  
	&\leq {c_8}\left\| {{{\bf{W}}^{k + 1}} - {{\bf{W}}^k}} \right\|_F^2,\\  
	\label{eq.h9}  
	\left\| {{\bm{\mu }}_2^{k + 1} - {\bm{\mu }}_2^k} \right\|_F^2 &= \left\| { - \lambda \nabla \phi \left( {{{\bf{E}}^{k + 1}}} \right) + \lambda \nabla \phi \left( {{{\bf{E}}^k}} \right)} \right\|_F^2 \nonumber \\  
	&\leq {c_9}\left\| {{{\bf{E}}^{k + 1}} - {{\bf{E}}^k}} \right\|_F^2, \\  
	\label{eq.h10}  
	\left\| {{\bm{\mu }}_3^{k + 1} - {\bm{\mu }}_3^k} \right\|_F^2 &= \left\| { - {\bf{F}}_\Omega ^{k + 1} + {\bf{F}}_\Omega ^k} \right\|_F^2 \nonumber \\  
	&\leq {c_{10}}\left\| {{\bf{F}}_\Omega ^{k + 1} - {\bf{F}}_\Omega ^k} \right\|_F^2.  
	\end{align}
	where $ {c_8} = {\alpha ^2}S_1^2 $ , $ {c_9} = {\lambda ^2}S_2^2 $ .
	
	This implies that
\begin{equation}\label{eq.h11}
\begin{array}{l}
L\left( {{{\bf{X}}^{k + 1}},\;{{\bf{W}}^{k + 1}},\;{{\cal E}^{k + 1}},\;{\bf{F}}_\Omega ^{k + 1},\;{{\bf{E}}^{k + 1}},\;{\bm{\mu }}_1^{k + 1},{\bm{\mu }}_2^{k + 1},{\bm{\mu }}_3^{k + 1}} \right)\;\\
\; - L\left( {{{\bf{X}}^{k + 1}},\;{{\bf{W}}^{k + 1}},\;{{\cal E}^{k + 1}},\;{\bf{F}}_\Omega ^{k + 1},\;{{\bf{E}}^k},\;{\bm{\mu }}_1^k,{\bm{\mu }}_2^k,{\bm{\mu }}_3^k} \right)\\
\; = \left\langle {{\bm{\mu }}_1^{k + 1} - {\bm{\mu }}_1^k,{{\bf{W}}^{k + 1}} - {{\bf{X}}^{k + 1}}} \right\rangle \\
\; + \left\langle {{\bm{\mu }}_2^{k + 1} - {\bm{\mu }}_2^k,{{\bf{E}}^{k + 1}} - {{\cal E}^{k + 1}}} \right\rangle \\
\; + \left\langle {{\bm{\mu }}_3^{k + 1} - {\bm{\mu }}_3^k,{{\bf{F}}_\Omega ^{k + 1}} - {{\bf{Y}}_\Omega } + {\bf{X}}_\Omega ^{k + 1} + {\cal E}_\Omega ^{k + 1}} \right\rangle \\
\; \le \frac{{{c_8}}}{\rho }\left\| {{{\bf{W}}^{k + 1}} - {{\bf{W}}^k}} \right\|_F^2 + \frac{{{c_9}}}{\rho }\left\| {{{\bf{E}}^{k + 1}} - {{\bf{E}}^k}} \right\|_F^2\\
\; + \frac{{{c_{10}}}}{\rho }\left\| {{\bf{F}}_\Omega ^{k + 1} - {\bf{F}}_\Omega ^k} \right\|_F^2.
\end{array}
\end{equation}

	Thus, Theorem \ref{th.7} is obtained.
\end{proof}	

\section{PROOF OF THEOREM 8}\label{app.8}
\begin{proof}
	Through Theorems \ref{th.4}-\ref{th.7}, the difference between two consecutive iterations of the function $ L  $ in the iterative process of Algorithm \ref{alg.1} is described as follows,
	\begin{equation}\label{eq.i1}
	\begin{array}{l}
	L\left( {{{\bf{X}}^{k + 1}},\;{{\bf{W}}^{k + 1}},\;{{\cal E}^{k + 1}},\;{\bf{F}}_\Omega ^{k + 1},\;{{\bf{E}}^{k + 1}},\;{\bm{\mu }}_1^{k + 1},{\bm{\mu }}_2^{k + 1},{\bm{\mu }}_3^{k + 1}} \right)\\
	\; - L\left( {{{\bf{X}}^k},\;{{\bf{W}}^k},\;{{\cal E}^k},\;{\bf{F}}_\Omega ^k,\;{{\bf{E}}^k},\;{\bm{\mu }}_1^k,{\bm{\mu }}_2^k,{\bm{\mu }}_3^k} \right)\\
	\; =  - {c_1}\left\| {{{\bf{X}}^{k + 1}} - {{\bf{X}}^k}} \right\|_F^2 - {c_2}\left\| {{\bf{X}}_\Omega ^{k + 1} - {\bf{X}}_\Omega ^k} \right\|_F^2\\
	\; - \left( {\alpha {c_3} - {c_8}} \right)\left\| {{{\bf{W}}^{k + 1}} - {{\bf{W}}^k}} \right\|_F^2 - {c_4}\left\| {{{\cal E}^{k + 1}} - {{\cal E}^k}} \right\|_F^2\\
	\; - {c_5}\left\| {{\cal E}_\Omega ^{k + 1} - {\cal E}_\Omega ^k} \right\|_F^2 - \left( {{c_6} - {c_{10}}} \right)\left\| {{\bf{F}}_\Omega ^{k + 1} - {\bf{F}}_\Omega ^k} \right\|_F^2\\
	\; - \left( {\lambda {c_7} - {c_9}} \right)\left\| {{{\bf{E}}^{k + 1}} - {{\bf{E}}^k}} \right\|_F^2
	\end{array}
	\end{equation}
	
	Then, the following inequality holds when $ \rho  \ge \max \left\{ {\frac{{{c_8}}}{{\alpha {c_3}}},\frac{{{c_9}}}{{\lambda {c_7}}},\frac{{{c_{10}}}}{{{c_6}}}} \right\} $,
	\begin{equation}\label{eq.i2}
	\begin{array}{l}
	L\left( {{{\bf{X}}^{k + 1}},\;{{\bf{W}}^{k + 1}},\;{{\cal E}^{k + 1}},\;{\bf{F}}_\Omega ^{k + 1},\;{{\bf{E}}^{k + 1}},\;{\bm{\mu }}_1^{k + 1},{\bm{\mu }}_2^{k + 1},{\bm{\mu }}_3^{k + 1}} \right)\;\\
	\; \le L\left( {{{\bf{X}}^k},\;{{\bf{W}}^k},\;{{\cal E}^k},\;{\bf{F}}_\Omega ^k,\;{{\bf{E}}^k},\;{\bm{\mu }}_1^k,{\bm{\mu }}_2^k,{\bm{\mu }}_3^k} \right).
	\end{array}
	\end{equation}
	Thus, Theorem \ref{th.8} is obtained.
\end{proof}	

\section{PROOF OF THEOREM 9}\label{app.9}
\begin{proof}
	According to \eqref{eq.b5}, the function $ L $ can be written in the following form,
	\begin{equation}\label{eq.j1}
	\begin{array}{l}
	L\left( {{{\bf{X}}^{k + 1}},\;{{\bf{W}}^{k + 1}},\;{{\cal E}^{k + 1}},\;{\bf{F}}_\Omega ^{k + 1},\;{{\bf{E}}^{k + 1}},\;{\bm{\mu }}_1^{k + 1},{\bm{\mu }}_2^{k + 1},{\bm{\mu }}_3^{k + 1}} \right)\\
	\; = \alpha R\left( {{{\bf{W}}^{k + 1}}} \right) + \lambda \phi \left( {{{\bf{E}}^{k + 1}}} \right) + \frac{1}{2}\left\| {{\bf{F}}_\Omega ^{k + 1}} \right\|_F^2\\
	\; + \frac{\rho }{2}\left\| {{{\bf{W}}^{k + 1}} - {{\bf{X}}^{k + 1}} + \frac{{{\bm{\mu }}_1^{k + 1}}}{\rho }} \right\|_F^2\\
	\; + \frac{\rho }{2}\left\| {{{\bf{E}}^{k + 1}} - {\cal E}_\Omega ^{k + 1} + \frac{{{\bm{\mu }}_2^{k + 1}}}{\rho }} \right\|_F^2\\
	\; + \frac{\rho }{2}\left\| {{\bf{F}}_\Omega ^{k + 1} - \left( {{{\bf{Y}}_\Omega } - {\bf{X}}_\Omega ^{k + 1} - {\cal E}_\Omega ^{k + 1}} \right) + \frac{{{\bm{\mu }}_3^{k + 1}}}{\rho }} \right\|_F^2\\
	\; \ge \alpha R\left( {{{\bf{W}}^{k + 1}}} \right) + \lambda \phi \left( {{{\bf{E}}^{k + 1}}} \right)\\
	\; + \frac{\rho }{2}\left\| {{{\bf{W}}^{k + 1}} - {{\bf{X}}^{k + 1}} + \frac{{{\bm{\mu }}_1^{k + 1}}}{\rho }} \right\|_F^2\\
	\; + \frac{\rho }{2}\left\| {{{\bf{E}}^{k + 1}} - {\cal E}_\Omega ^{k + 1} + \frac{{{\bm{\mu }}_2^{k + 1}}}{\rho }} \right\|_F^2\\
	\; + \frac{\rho }{2}\left\| {{\bf{F}}_\Omega ^{k + 1} - \left( {{{\bf{Y}}_\Omega } - {\bf{X}}_\Omega ^{k + 1} - {\cal E}_\Omega ^{k + 1}} \right) + \frac{{{\bm{\mu }}_3^{k + 1}}}{\rho }} \right\|_F^2.
	\end{array}
	\end{equation}
	
	Using the property of inequality $ {\left( {a + b} \right)^2} \le 2\left( {{a^2} + {b^2}} \right) $ with $ a,b \ge 0 $, the lower bound of each term can be estimated, 	
	\begin{equation}\label{eq.j2}
	\begin{array}{l}
	\left\| {{{\bf{W}}^{k + 1}} - {{\bf{X}}^{k + 1}} + \frac{{{\bm{\mu }}_1^{k + 1}}}{\rho }} \right\|_F^2\;\\
	\; \ge \frac{1}{2}\left\| {{{\bf{W}}^{k + 1}}} \right\|_F^2\;\; - \left\| {{{\bf{X}}^{k + 1}} - \frac{{{\bm{\mu }}_1^{k + 1}}}{\rho }} \right\|_F^2,
	\end{array}
	\end{equation}	
	\begin{equation}\label{eq.j3}
	\begin{array}{l}
	\left\| {{{\bf{E}}^{k + 1}} - {\cal E}_\Omega ^{k + 1} + \frac{{{\bm{\mu }}_2^{k + 1}}}{\rho }} \right\|_F^2\\
	\; \ge \frac{1}{2}\left\| {{{\bf{E}}^{k + 1}}} \right\|_F^2\; - \left\| {{\cal E}_\Omega ^{k + 1} - \frac{{{\bm{\mu }}_2^{k + 1}}}{\rho }} \right\|_F^2,
	\end{array}
	\end{equation}
	\begin{equation}\label{eq.j4}
	\begin{array}{l}
	\left\| {{\bf{F}}_\Omega ^{k + 1} - \left( {{{\bf{Y}}_\Omega } - {\bf{X}}_\Omega ^{k + 1} - {\cal E}_\Omega ^{k + 1}} \right) + \frac{{{\bm{\mu }}_3^{k + 1}}}{\rho }} \right\|_F^2\\
	\; \ge \frac{1}{2}\left\| {{\bf{F}}_\Omega ^{k + 1}} \right\|_F^2 - \left\| {{{\bf{Y}}_\Omega } - {\bf{X}}_\Omega ^{k + 1} - {\cal E}_\Omega ^{k + 1} - \frac{{{\bm{\mu }}_3^{k + 1}}}{\rho }} \right\|_F^2.
	\end{array}
	\end{equation}

	Furthermore, it follows from Assumption \ref{Ass.2} that $ R\left( {{{\bf{W}}^{k + 1}}} \right) $ and $ \phi \left( {{{\bf{E}}^{k + 1}}} \right) $ are bounded.
	Therefore, we can conclude that the sequence $ \left\{ {{{\bf{\Gamma }}^k}} \right\}_{k = 1}^{ + \infty } $ is bounded.
	
	Thus, Theorem \ref{th.9} is obtained.
\end{proof}	

\section{PROOF OF THEOREM 10}\label{app.10}
\begin{proof}
	Theorems \ref{th.4}-\ref{th.7}  have demonstrated that the function $ L $ monotonically decreases and possesses a lower bound, which implies the convergence of the function  $ L $. 
	The following conclusions can be derived from \eqref{eq.i1}.
	\[\begin{array}{l}
	\mathop {\lim }\limits_{k \to  + \infty } \left\| {{{\bf{X}}^{k + 1}} - {{\bf{X}}^k}} \right\|_F^2 = 0,\;\;\;\mathop {\lim }\limits_{k \to  + \infty } \left\| {{\bf{X}}_\Omega ^{k + 1} - {\bf{X}}_\Omega ^k} \right\|_F^2 = 0,\\
	\mathop {\lim }\limits_{k \to  + \infty } \left\| {{{\bf{W}}^{k + 1}} - {{\bf{W}}^k}} \right\|_F^2 = 0,\;\mathop {\lim }\limits_{k \to  + \infty } \left\| {{{\cal E}^{k + 1}} - {{\cal E}^k}} \right\|_F^2 = 0,\\
	\mathop {\lim }\limits_{k \to  + \infty } \left\| {{\cal E}_\Omega ^{k + 1} - {\cal E}_\Omega ^k} \right\|_F^2 = 0,\;\;\;\;\;\mathop {\lim }\limits_{k \to  + \infty } \left\| {{\bf{F}}_\Omega ^{k + 1} - {\bf{F}}_\Omega ^k} \right\|_F^2 = 0,\\
	\mathop {\lim }\limits_{k \to  + \infty } \left\| {{{\bf{E}}^{k + 1}} - {{\bf{E}}^k}} \right\|_F^2 = 0,\;\;\;\;\mathop {\lim }\limits_{k \to  + \infty } \left\| {{\bm{\mu }}_1^{k + 1} - {\bm{\mu }}_1^k} \right\|_F^2 = 0,\\
	\mathop {\lim }\limits_{k \to  + \infty } \left\| {{\bm{\mu }}_2^{k + 1} - {\bm{\mu }}_2^k} \right\|_F^2 = 0,\;\;\;\;\;\mathop {\lim }\limits_{k \to  + \infty } \left\| {{\bm{\mu }}_3^{k + 1} - {\bm{\mu }}_3^k} \right\|_F^2 = 0.
	\end{array}\] 
	Therefore, we can conclude that $ \mathop {\lim }\limits_{k \to  + \infty } \left\| {{{\bf{\Gamma }}^{k + 1}} - {{\bf{\Gamma }}^k}} \right\|_F^2 = 0 $. 

	Thus, Theorem \ref{th.10} is obtained.
\end{proof}	

\section{PROOF OF THEOREM 11}\label{app.11}
\begin{proof}
	Theorems \ref{th.4}-\ref{th.10} have demonstrated that the sequence $ \left\{ {{{\bf{\Gamma }}^k}} \right\}_{k = 1}^{ + \infty } $ is bounded and satisfies the condition $ \mathop {\lim }\limits_{k \to  + \infty } \left\| {{{\bf{\Gamma }}^{k + 1}} - {{\bf{\Gamma }}^k}} \right\|_F^2 = 0 $. 
	According to the Bolzano-Weierstrass theorem \cite{jiao2016alternating}, we can deduce that there must exist a subsequence $ \left\{ {{{\bf{\Gamma }}^{{k_j}}}} \right\} $ of the sequence $ \left\{ {{{\bf{\Gamma }}^k}} \right\} $, such that $ \mathop {\lim }\limits_{k \to  + \infty } {{\bf{\Gamma }}^{{k_j}}} = {{\bf{\Gamma }}^ * } $.
	
	Next, considering the sufficient condition for the minimization problem in Algorithm \ref{alg.1}, which requires the following equation to be satisfied,
	\[\begin{array}{l}
	{{\bf{W}}^k} - {{\bf{X}}^{k + 1}} + \frac{{{\bm{\mu }}_1^k}}{\rho } + {\bf{F}}_\Omega ^k - \left( {{{\bf{Y}}_\Omega } - {\bf{X}}_\Omega ^{k + 1} - {\cal E}_\Omega ^k} \right) + \frac{{{\bm{\mu }}_3^k}}{\rho } = 0,\\
	\alpha \nabla R\left( {{{\bf{W}}^{k + 1}}} \right) + \rho \left( {{{\bf{W}}^{k + 1}} - {{\bf{X}}^{k + 1}} + \frac{{{\bm{\mu }}_1^k}}{\rho }} \right) = 0,\\
	{{\bf{E}}^k} - {{\cal E}^{k + 1}} + \frac{{{\bm{\mu }}_2^k}}{\rho } + {\bf{F}}_\Omega ^k - \left( {{{\bf{Y}}_\Omega } - {\bf{X}}_\Omega ^{k + 1} - {\cal E}_\Omega ^k} \right) + \frac{{{\bm{\mu }}_3^k}}{\rho } = 0,\\
	{\bf{F}}_\Omega ^{k + 1} + \rho \left( {{\bf{F}}_\Omega ^{k + 1} - \left( {{{\bf{Y}}_\Omega } - {\bf{X}}_\Omega ^{k + 1} - {\cal E}_\Omega ^{k + 1}} \right) + \frac{{{\bm{\mu }}_3^k}}{\rho }} \right) = 0,\\
	\lambda \nabla \phi \left( {{{\bf{E}}^{k + 1}}} \right) + \rho \left( {{{\bf{E}}^{k + 1}} - {{\cal E}^{k + 1}} + \frac{{{\bm{\mu }}_2^k}}{\rho }} \right) = 0.
	\end{array}\]
	
	Finally, by taking limits of the above five equations along the subsequence $ \left\{ {{{\bf{\Gamma }}^{{k_j}}}} \right\} $, we obtain
	\[\begin{array}{l}
	{\bm{\mu }}_1^ *  =  - \alpha \nabla R\left( {{{\bf{W}}^ * }} \right),\;\;\;\,{\bm{\mu }}_2^ *  =  - \lambda \nabla \phi \left( {{{\bf{E}}^ * }} \right),\;{\bm{\mu }}_3^ *  =  - {\bf{F}}_\Omega ^ * 	\vspace{2mm}, \\
	{{\bf{W}}^ * } = {{\bf{X}}^ * },\;{{\bf{E}}^ * } = {{\cal E}^ * },\;\;{\bf{F}}_\Omega ^ *  = {{\bf{Y}}_\Omega } - {\bf{X}}_\Omega ^ *  - {\cal E}_\Omega ^ *. \\
	\end{array}\]

	Therefore, we have demonstrated that the limit point $ {{\bf{\Gamma }}^ * } $ is a stable point of formula \eqref{eq.b5}.
	
	Thus, Theorem \ref{th.11} is obtained.
\end{proof}


%







\end{document}